%% file: paper.tex
\documentclass[10pt,journal,cspaper,final,twocolumn,compsoc]{./IEEEtran}
\pdfoutput=1
\usepackage[pdftex]{graphicx}
\usepackage[cmex10]{amsmath} 

\usepackage{float}
\usepackage{bm}
\usepackage{adjustbox}     
\usepackage{enumitem}
\usepackage{colortbl}

\input{macros}
\graphicspath{
{figures/}
{images/}
}
\def\iid{iid\xspace}

\def\x{\mathbf{x}}

\def\u{\mathbf{u}}

\def\B{\text{B}}
\DeclareMathOperator*{\argmax}{arg\,max}

\makeatletter

\floatstyle{ruled}
\newfloat{algorithm}{tbp}{loa}
\floatname{algorithm}{Algorithm}

\ifCLASSOPTIONcompsoc
   \usepackage[nocompress]{cite} 
\else
  \usepackage{cite}
\fi
\usepackage{algorithmic} 
\usepackage{array} 

\ifCLASSOPTIONcompsoc
  \usepackage[caption=false,font=footnotesize,labelfont=sf,textfont=sf]{subfig} 
\else
  \usepackage[caption=false,font=footnotesize]{subfig}
\fi

\usepackage{url} 

\usepackage[pagebackref=false,breaklinks=true,colorlinks,bookmarks=false]{hyperref} 

\makeatother

\def\VOC{PASCAL VOC }
\def\iid{iid\xspace}
\def\rev[#1][#2]{{\color{blue} \sout{#1}} {\bf \color{red} #2}} 

\begin{document}

\title{Approximate Fisher Kernels of non-iid Image Models for Image Categorization}

\author{Ramazan~Gokberk~Cinbis, %
        Jakob~Verbeek, %
        and~Cordelia~Schmid,~\IEEEmembership{Fellow,~IEEE}%
\IEEEcompsocitemizethanks{\IEEEcompsocthanksitem R. G. Cinbis is with Milsoft, Ankara, Turkey. Most of the work in this paper was done when he was with the LEAR team, Inria Grenoble, France. E-mail: firstname.lastname@inria.fr\protect\\
\IEEEcompsocthanksitem J. Verbeek and C. Schmid are with LEAR team, Inria Grenoble Rh\^one-Alpes, Laboratoire Jean Kuntzmann, CNRS, Univ.\ Grenoble Alpes, France. E-mail: firstname.lastname@inria.fr}
\thanks{Copyright (c) 2015 IEEE. Personal use of this material is permitted. However, permission to use this material for any other purposes must be obtained from the IEEE by sending a request to pubs-permissions@ieee.org.}}

\markboth{to appear in IEEE Transactions on Pattern Analysis and Machine Intelligence, 2015}
{}
\IEEEtitleabstractindextext{%
\input{abstract}

\begin{IEEEkeywords}
Statistical image representations, object recognition, image classification, Fisher kernels.
\end{IEEEkeywords}
}
\maketitle
\IEEEdisplaynontitleabstractindextext

\def\mysingleplot#1{\includegraphics[width=.7\linewidth]{#1}}

\def\mydoubleplot#1{\includegraphics[width=.49\linewidth]{#1}} 

\input{introduction}

\input{related}

\input{fisher}

\input{model}

\input{experiments}

\input{conclusion}

\input{appendices}

\bibliographystyle{ieee}
\bibliography{bibabbr,jjv.copy,rgc}
\vspace{-0.3mm}
\input{bios}

\end{document}

%% file: macros.tex
\newif\ifrevfinal
\revfinalfalse
\def\rev[#1][#2]{\ifrevfinal #2 \else {\color{blue} \sout{#1}} {\bf \color{red} #2} \fi}
\usepackage{color}
\usepackage[normalem]{ulem}

\def\etal{et al.\ }

\def\grad#1#2{\frac{\partial #1}{\partial #2}}  
\def\ex#1#2{\textrm{I\!E}_{#1}\!\left[#2\right]}     
\def\half{\frac{1}{2}}              
\def\<{\langle}
\def\>{\rangle}

\def\Eq#1{Eq.~(\ref{eq:#1})}
\def\eq#1{(\ref{eq:#1})}
\def\be{\begin{equation}}
\def\ee{\end{equation}}
\def\bea{\begin{eqnarray}}
\def\eea{\end{eqnarray}}
\def\fig#1{Figure~\ref{fig:#1}}
\def\tab#1{Table~\ref{tab:#1}}
\def\sect#1{Section~\ref{sec:#1}}

\def\app#1{Appendix~\ref{app:#1}}

\makeatletter
\gdef\SetFigFont#1#2#3{%
  \reset@font\fontsize{10}{12pt}%
  \selectfont%
}
\let\eqnarray@=\eqnarray \let\endeqnarray@=\endeqnarray
\def\eqnarray{\bgroup\arraycolsep=2pt\eqnarray@}
\def\endeqnarray{\endeqnarray@\egroup}
\makeatother

\def\em{\slshape}

\usepackage{times,amsmath,amssymb,amsbsy,xspace}
\makeatletter
\DeclareRobustCommand\onedot{\futurelet\@let@token\@onedot}
\def\@onedot{\ifx\@let@token.\else.\null\fi\xspace}

\def\eg{\emph{e.g}\onedot} 
\def\ie{\emph{i.e}\onedot} 
\def\cf{\emph{c.f}\onedot} 
 \def\vs{\emph{vs}\onedot} 
\def\wrt{w.r.t\onedot} 
\def\etal{\emph{et al}\onedot}

\newcounter{iictr}
\def\@iia[#1]{\setcounter{iictr}{#1}\@iib}
\def\@iib{\iii[\roman{iictr}]\xspace}
\def\ii{\@ifnextchar[{\@iia}{\stepcounter{iictr}\@iib}}
\def\iii[#1]{({\em#1\/})}
\makeatother

%% file: abstract.tex
\begin{abstract} 
The bag-of-words (BoW) model treats images as sets of local descriptors and represents them by visual word histograms. 
The Fisher vector (FV) representation extends BoW, by considering the first and second order statistics of local descriptors.
In both representations local descriptors are assumed to be identically and independently distributed (\iid), which is a poor assumption from a modeling  perspective.  
It has been experimentally observed that the performance of BoW and FV representations can be improved by employing discounting transformations such as power normalization. 
In this paper, we introduce non-\iid models by treating the model parameters as latent variables which are integrated out, rendering all local regions
dependent.  
Using the Fisher kernel principle we encode an image by the gradient of the data log-likelihood \wrt the model hyper-parameters.  
Our models naturally generate discounting effects in the representations; suggesting that  such transformations have proven successful because they closely correspond to the representations obtained for non-iid models. 
To enable tractable computation, we rely on variational free-energy bounds to learn the hyper-parameters and to compute approximate Fisher kernels.   
Our experimental evaluation results validate that our models lead to performance
improvements comparable to using power normalization, as  employed in state-of-the-art feature aggregation methods.
\end{abstract}

%% file: introduction.tex
\section{Introduction}
\label{sec:intro}

\begin{figure}
\begin{center}
\mysingleplot{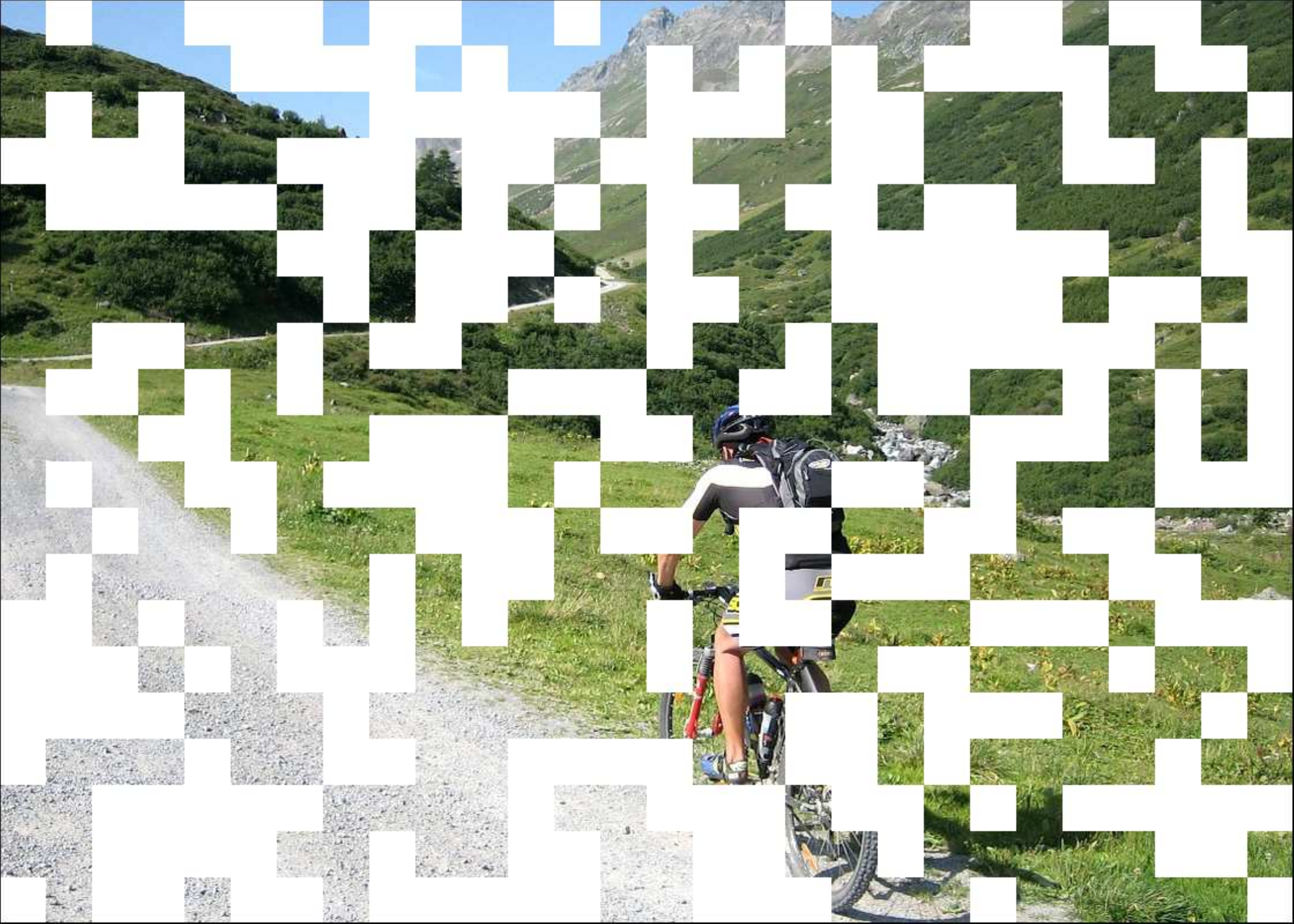}
\end{center}
\caption{Local image appearance is not \iid: the visible regions
are informative on the masked-out ones; one has the impression to have
seen the complete image by looking at half of the pixels.   
}
\label{fig:iid}
\end{figure}

\IEEEPARstart{P}{atch}-based
image representations, such bag of visual
words (BoW) \cite{dance04eccv, sivic03iccv}, are widely
utilized in image categorization and retrieval systems.
BoW descriptor
represents an image as a histogram over visual word
counts. The histograms are constructed by mapping local
feature vectors in images to cluster indices, where the
clustering is typically learned using k-means.
Perronnin and Dance \cite{perronnin07cvpr} have enhanced this basic
representation using the notion of Fisher kernels
\cite{jaakkola99nips}.  In this case local descriptors
are soft-assigned to components of a mixture of Gaussian
(MoG) density, and the image is represented using the
gradient of the log-likelihood of the local descriptors
\wrt the MoG parameters.  As we show below, both BoW as
well as MoG Fisher vector representations are based on
models that assume that local descriptors are
independently and identically distributed (\iid). However,
the \iid assumption is a very poor one from a modeling
perspective, see the illustration in \fig{iid}. 
 
In this work, we consider models that capture the
dependencies among local image regions by means of
non-\iid but completely exchangeable models, \ie like \iid
models our models still treat the image as an unordered
set of regions.  We treat the parameters of the BoW models
as latent variables with prior distributions learned from
data. By integrating out the latent variables, all image
regions become mutually dependent.  We generate image
representations from these models by applying the Fisher
kernel principle, in this case by taking the gradient of
the log-likelihood of the data in an image \wrt the
hyper-parameters that control the priors on the latent
model parameters. 

However, in some cases, the gradient of the log-likelihood
of the data can be intractable to compute. 
To compute a gradient-based representation in such cases, we replace the intractable log-likelihood with a tractable variational bound. We then compute gradients with respect to this bound instead of the likelihood.
Following
\cite{chandalia06nipsw}, which is the first and one of the
very few studies utilizing this approximation method, we
refer to the resulting kernel as the {\em variational
Fisher kernel}. We show that the variational Fisher kernel is equivalent to the actual
Fisher kernel when the variational bound is tight.
Therefore, the variational Fisher kernel provides not only
a technique for approximating intractable Fisher kernels,
but also an alternative formulation for computing exact
Fisher kernels. We  demonstrate through examples that the
variational formulation can be mathematically more convenient
for deriving Fisher vectors representations.

In this work, we analyze three non-\iid image models.  Our
first model is the multivariate P\'{o}lya model which
represents the set of visual word indices of an image as
independent draws from an unobserved multinomial
distribution, itself drawn from a Dirichlet prior
distribution. By integrating out the latent multinomial
distribution, a model is obtained in which all visual word
indices are mutually dependent.  Interestingly, we find
that our non-\iid models yield gradients  that are
qualitatively similar to popular transformations of
BoW image representations, such as square-rooting
histogram entries or more generally applying power normalization
\cite{jegou11pami,perronnin10cvpr1,perronnin10eccv,vedaldi10cvpr}.
Therefore, our first contribution is to show that such
transformations appear naturally if  we remove  the unrealistic
\iid assumption, \ie, to provide an explanation why such
transformations are  beneficial.  

Our second contribution is the analysis of Fisher vector
representations over the latent Dirichlet allocation (LDA)
model \cite{blei03jmlr} for image classification purposes.
The LDA model can capture the co-occurrence statistics
missing in BoW representations.  In this case the
computation of the gradients is intractable, therefore, we
compute approximate variational Fisher vectors
\cite{chandalia06nipsw}.  We compare performance to Fisher
vectors of PLSA \cite{hofmann01ml}, a topic model that
does not treat the model parameters as latent variables.
We find that topic models improve over BoW models, and
that the LDA  improves over PLSA even when square-rooting
is applied.

Our third  contribution is our most advanced model, which
assumes that the local descriptors are \iid samples from a
latent MoG distribution, and we integrate out the mixing
weights, means and variances of the MoG distribution.
Since the computation of the gradients is intractable, we
also use the variational Fisher kernel framework for this model.
  This leads to a representation that performs on
par with the improved Fisher vector (FV) representation of
\cite{perronnin10eccv} based on \iid MoG models, which
includes power normalization. 

In our experimental analysis, we present a detailed
experimental evaluation of the proposed the non-\iid image
models over  local SIFT descriptors.  In addition, we
demonstrate that the latent MoG image model can
effectively be combined with Convolutional Neural Network
(CNN) based features. 
 We consider two approaches for this purpose. First, following recent work \cite{gong14eccv,liu14nips}, we compute Fisher vectors over densely sampled image patches that are encoded using CNN features. Second, we propose to extract Fisher vectors over image regions sampled by a selective search method \cite{uijlings13ijcv}.  
The experimental results on the
\VOC 2007~\cite{everingham15ijcv} and MIT Indoor
Scenes~\cite{quattoni09cvpr} datasets confirm the
effectiveness of the proposed latent MoG image model, and
the corresponding non-\iid image descriptors.

This paper extends our earlier paper \cite{cinbis12cvpr}.
We present more complete and detailed discussions of related work and the variational Fisher kernel framework. We give a proof that Fisher kernels given by the traditional form and the variational framework are equivalent when the variational bound is tight. 
We extend the experimental evaluation of the proposed non-\iid MoG
models by evaluating them over CNN-based local
descriptors.  We also show that the classification results
can be further improved by computing the CNN features over
selective search windows, compared to using densely
sampled image regions. We perform additional
experimental evaluation on the MIT Indoor
dataset~\cite{quattoni09cvpr}. Finally, we present
a new empirical study on the relationship
between the model likelihood and image categorization
performance.

%% file: related.tex
\section{Related Work}
\label{sec:related}

The use of non-linear feature transformations in BoW image
representations is widely recognized to be beneficial for
image categorization
\cite{jegou11pami,perronnin10cvpr1,perronnin10eccv,vedaldi10cvpr,zhang07ijcv}.
These transformations alleviate an obvious shortcoming of
linear classifiers on BoW image representations: the fact
that a fixed change $\Delta$ in a BoW histogram, from $h$
to $h+\Delta$, leads to a  score increment that is
independent of the  original histogram $h$: $f(h+\Delta) -
f(h) = w^\top(h+\Delta) - w^\top h = w^\top \Delta$.
This means that the effect on the score for a change $\Delta$ is not dependent on the context $h$ in which it appears.
Therefore, the score increment from images (a) though  (d)
in \fig{cows} will be  comparable, which is
undesirable: the classifier  score for \emph{cow} should
sharply increase from (a) to (b), and then remain stable
among (b), (c), and (d).  

\def\mypicwithcap#1#2{%
    \subfloat[#2]{%
        \includegraphics[height=.3\linewidth,width=.43\linewidth]{#1}
    }}
\begin{figure}
\begin{center}
\mypicwithcap{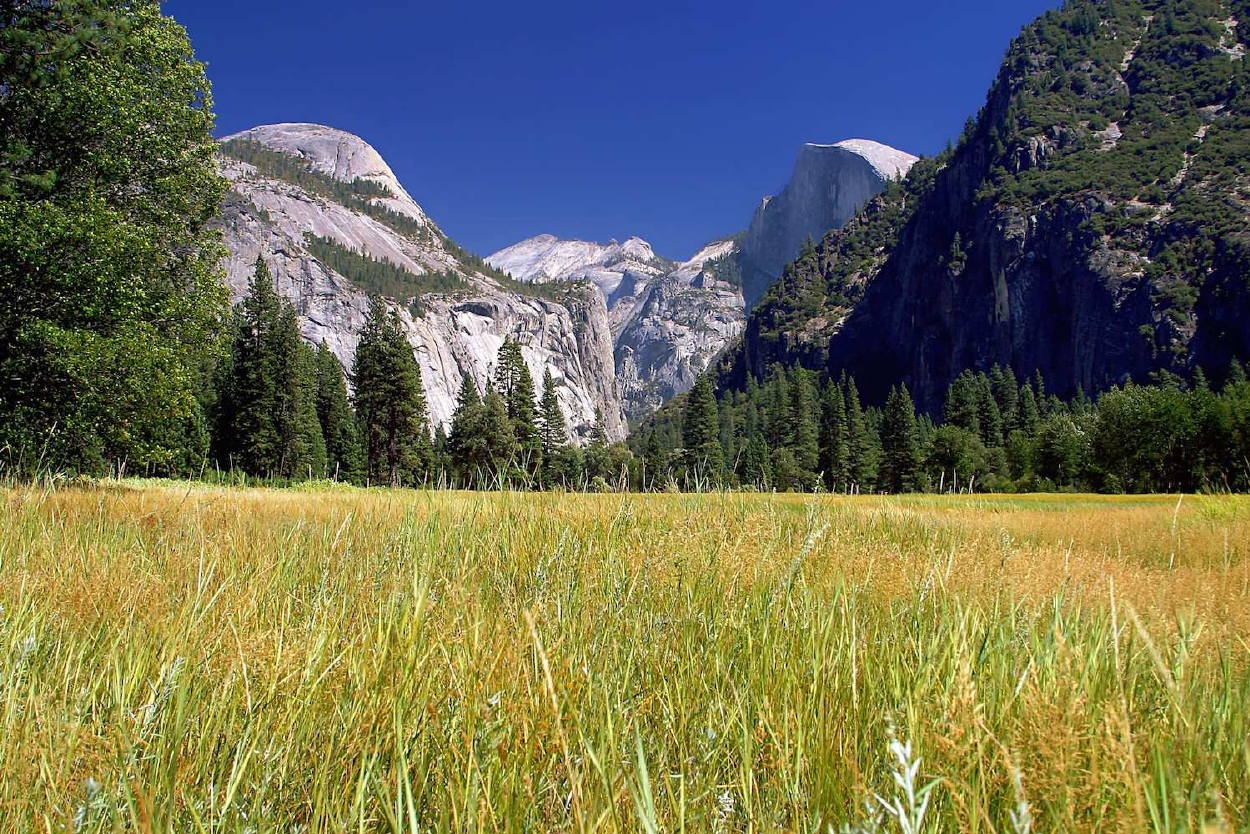}{}
\mypicwithcap{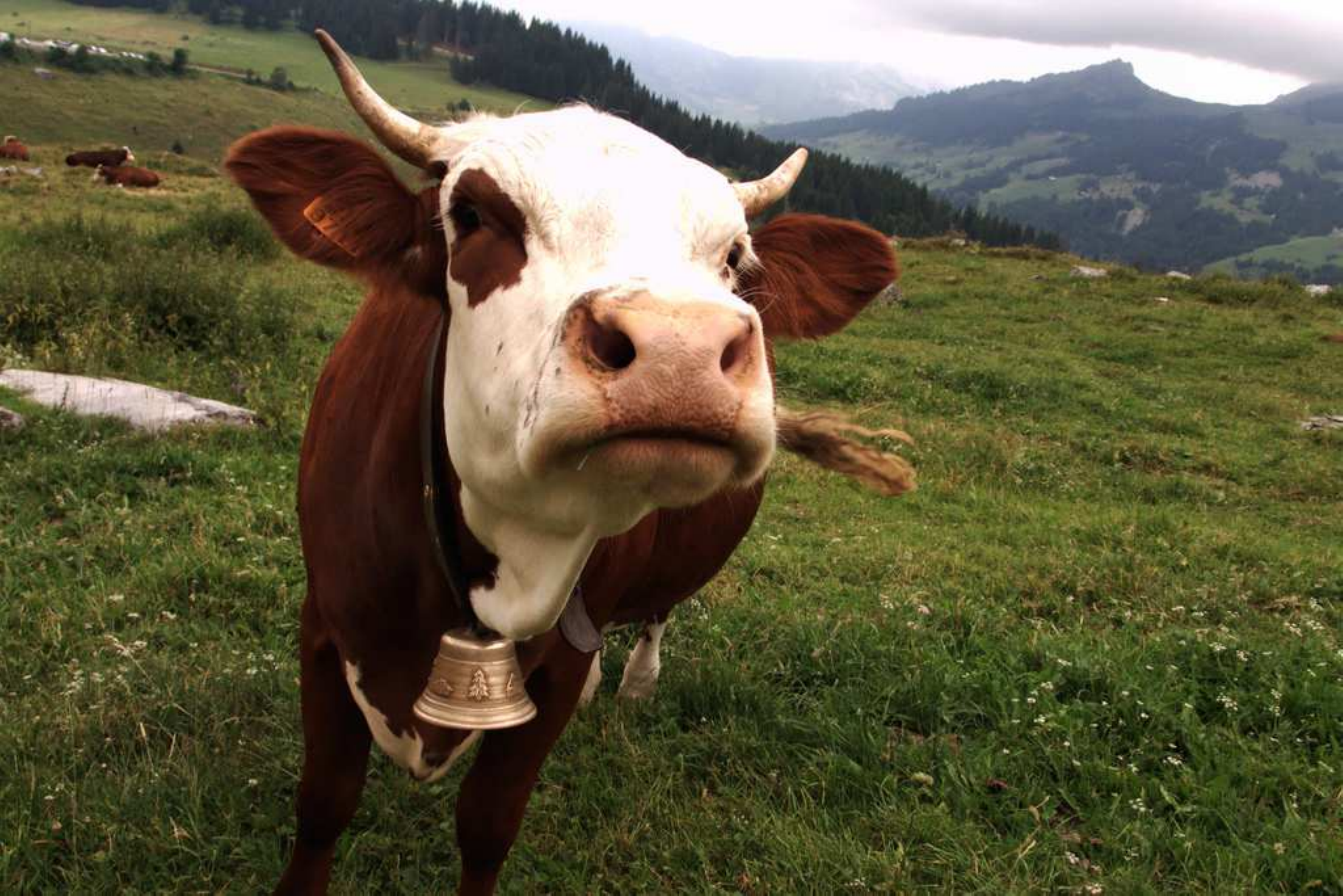}{} \\[-2mm]
\mypicwithcap{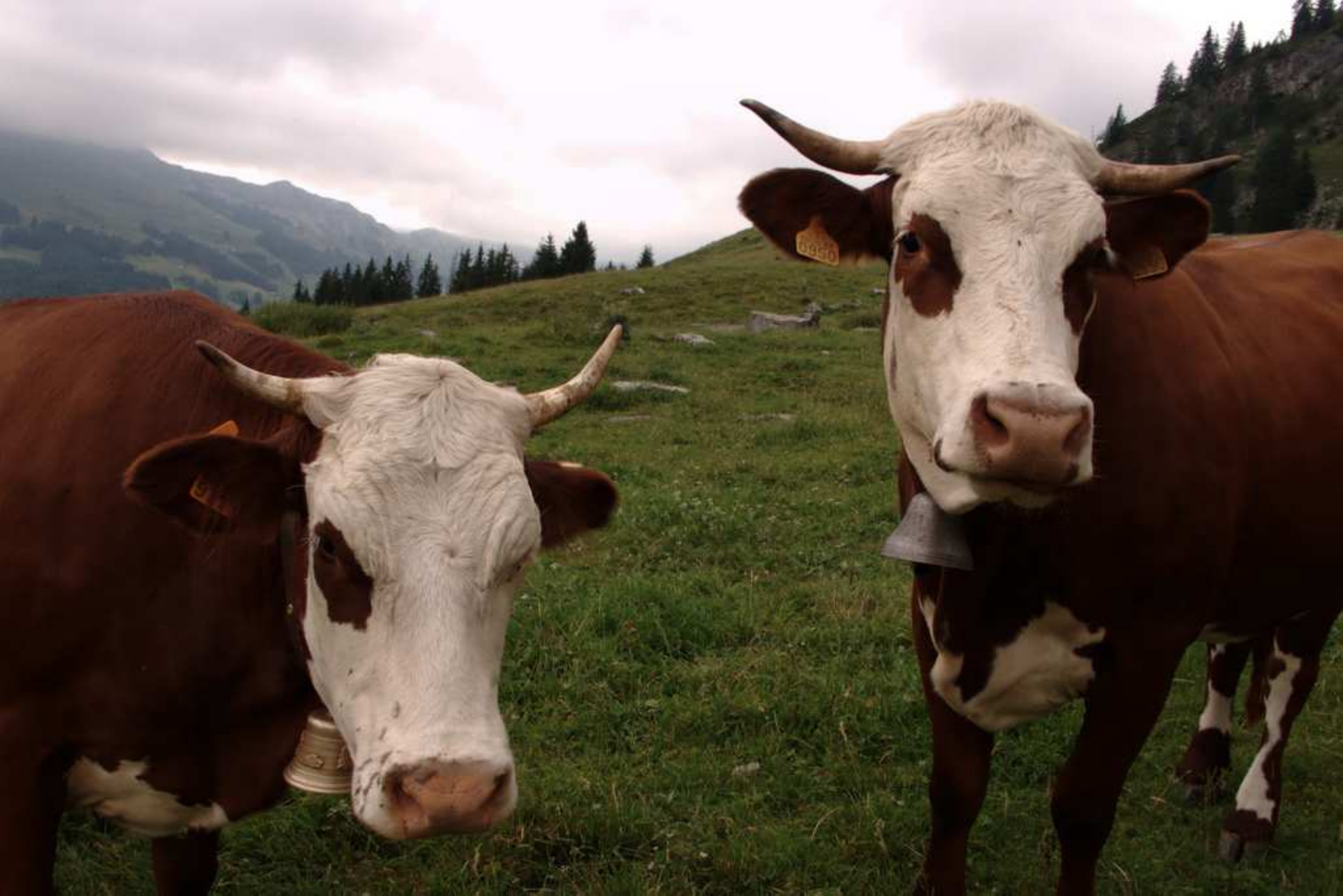}{}
\mypicwithcap{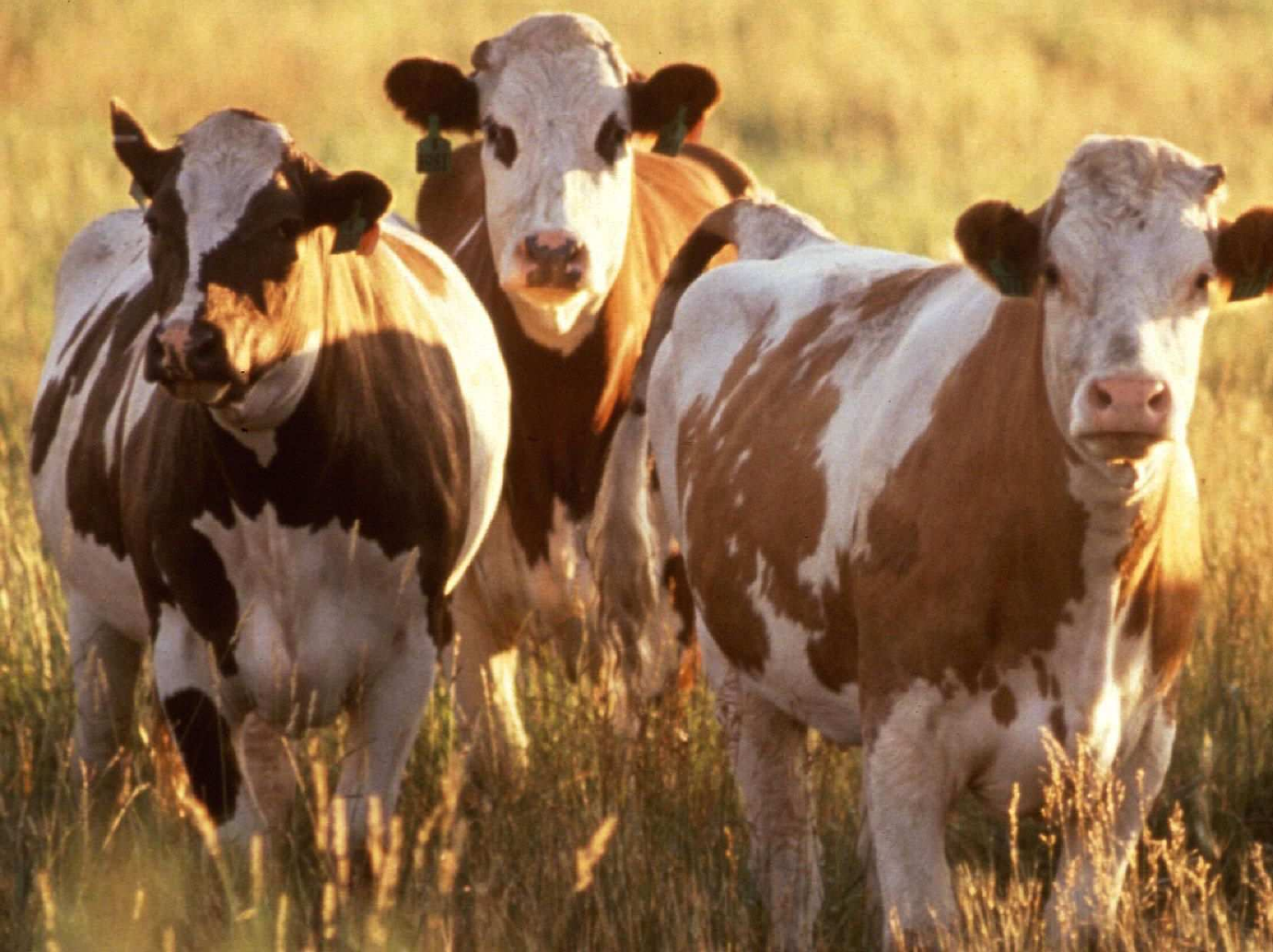}{}
\end{center}
\caption{
The score of a linear `cow' classifier will increase similarly from images (a)
through (d) due to the increasing number of cow patches. This is
undesirable:  the  score  should sharply
increase from (a) to (b), and remain stable among (b),  (c), and
(d).    
}
\label{fig:cows}
\end{figure}

Popular remedies to this problem include  the use of
chi-square kernels \cite{zhang07ijcv}, or taking the
square-root of histogram entries \cite{perronnin10cvpr1},
also referred to as the Hellinger kernel
\cite{vedaldi10cvpr}. Power normalization
\cite{perronnin10cvpr1}, defined as $f(x)=sign(x)|x|^\rho$, is
a similar transformation that can be applied to
non-histogram feature vectors, and it is equivalent to
signed square-rooting for the
coefficient $\rho=1/2$.  The effect of all of these is similar:
they transform the features such that the first few
occurrences of visual words will have a more pronounced
effect on the classifier score than if the count is
increased by the same amount but starting at a larger
value.  This is desirable, since now the first patches
providing evidence for an object category can
significantly impact the score, and hence making it for example easier to
detect small object instances. The qualitative similarity
is illustrated in \fig{dists}, where we compare the
$\ell_2$, chi-square, and Hellinger distances on the range
$[0,1]$. 

The motivation for these transformations  tends to vary in the literature. 
Sometimes it is based on
empirical observations of improved performance
\cite{perronnin10cvpr1,vedaldi10cvpr}, by reducing
sparsity in Fisher vectors \cite{perronnin10eccv}, or in
terms of variance stabilizing
transformations~\cite{jegou11pami,winn05iccv}. 
Recently, Kobayashi \cite{kobayashi14cvpr} showed that a
similar discounting transformation based on taking
logarithm of histogram entries, can be derived via
modeling $\ell_1$-normalized descriptors by Dirichlet
distribution. Rana \etal~\cite{rana14accvw} propose to
discriminatively learn power normalization coefficients
for image retrieval using a triplet-based objective
function, which aims to obtain smaller distances across
matching image pairs than non-matching ones.  In contrast
to these studies, we show that such discounting
transformations appear naturally in generative image
models that avoid making the unrealistic \iid assumption that underlies the standard BoW and MoG-FV image representations.  

Similar transformations are also used in image retrieval
to counter burstiness effects~\cite{jegou09cvpr}, \ie, if
rare visual words occur in an image, they tend to do so in
bursts due to the locally repetitive nature of natural
images. Burstiness also occurs in text, and the Dirichlet
compound multinomial distribution, also known as
multivariate P\'{o}lya distribution, has been used to
model this effect~\cite{madsen05icml}.  This model places a
Dirichlet prior on a latent per-document multinomial, and
words in a document are sampled independently from it.
Elkan \cite{elkan05spire} shows the relationship between
the Fisher kernel of the multivariate P\'{o}lya
distribution and the tf-idf document representation. In
\sect{model}, we investigate the Fisher kernel
based on multivariate P\'{o}lya distribution as our most
basic non-\iid image representation.

\def\mypicwithcap#1#2{%
    \subfloat[#2]{%
    \includegraphics[height=.3\linewidth,width=.3\linewidth]{#1}
    }}
\begin{figure}
\begin{center}
\mypicwithcap{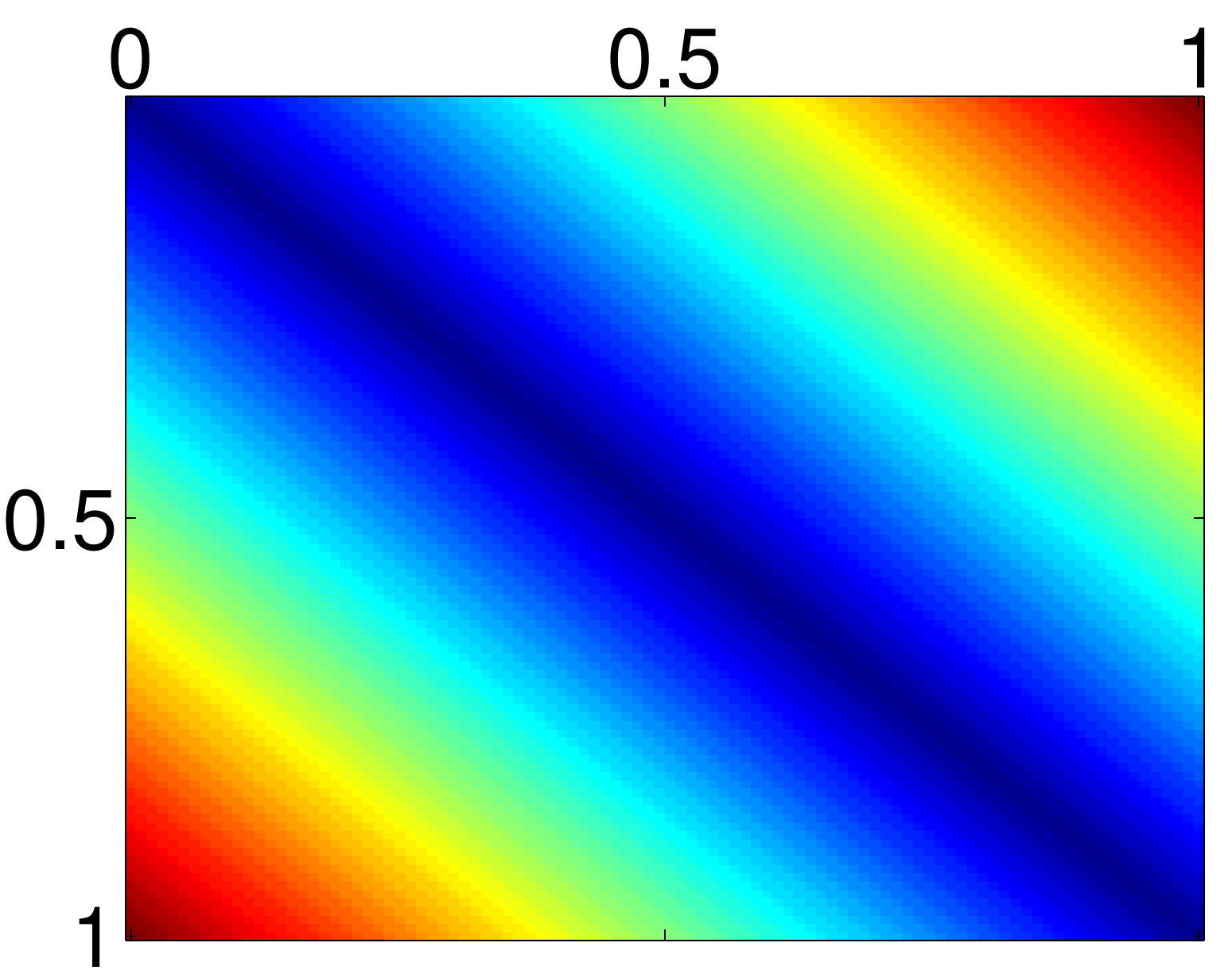}{$\ell_2$}
\mypicwithcap{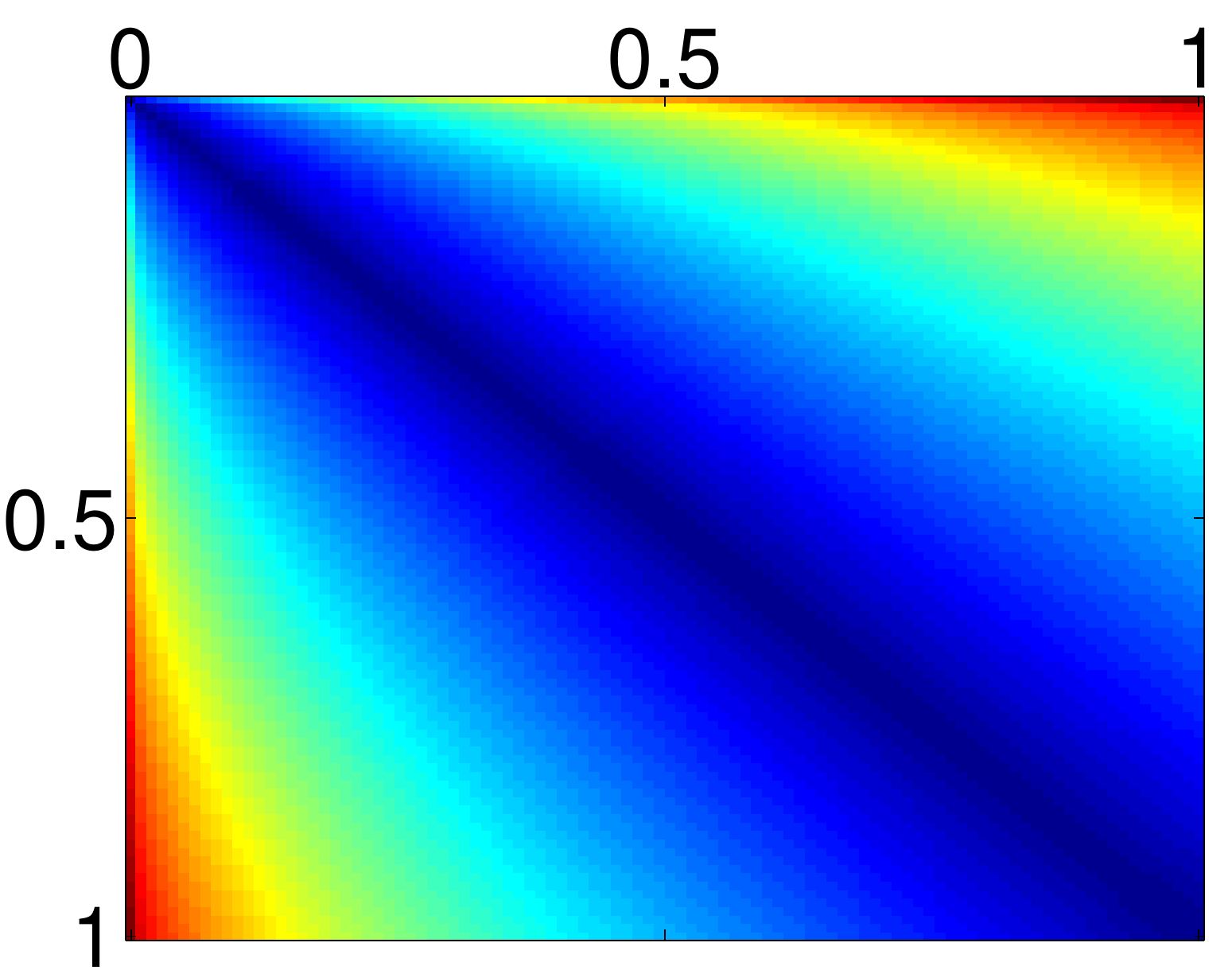}{Hellinger}
\mypicwithcap{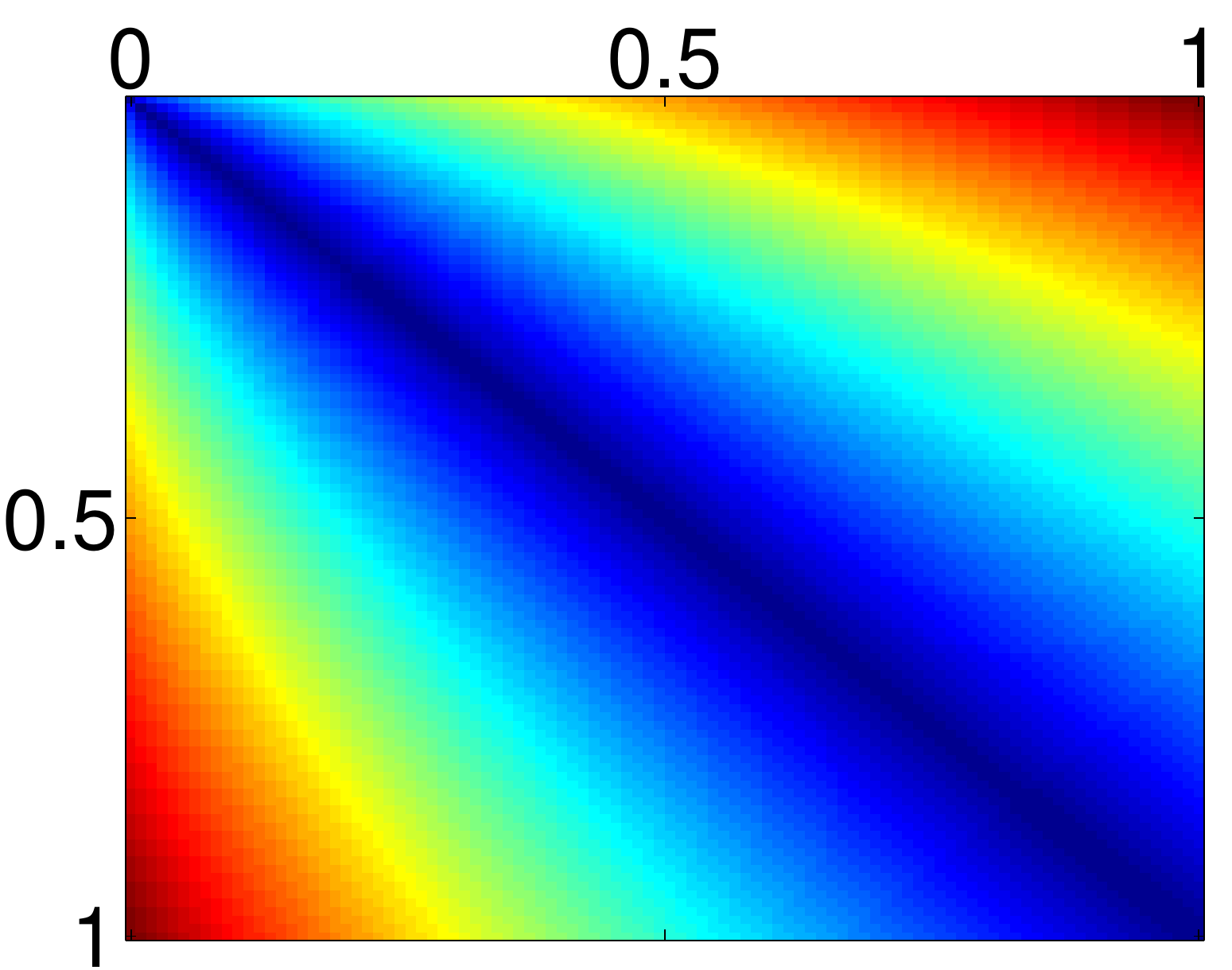}{chi-square}
\end{center}
\caption{Comparison of  $\ell_2$,
Hellinger, and chi-square distances for values in the unit interval.
Both the Hellinger and chi-square
distance discount the effect of small changes in large
values, unlike the $\ell_2$ distance.}
\label{fig:dists}
\end{figure}

Our use of latent Dirichlet allocation (LDA)
\cite{blei03jmlr} differs from earlier work on using topic
models such as LDA or PLSA \cite{hofmann01ml} for object
recognition~\cite{larlus09ivc,quelhas05iccv}. The latter
use topic models to compress BoW image
representations by using the inferred document-specific
topic distribution. Similarly, Chandalia and
Beal~\cite{chandalia06nipsw} propose to compress BoW
document representation by computing LDA Fisher vector
with respect to the parameters of the Dirichlet prior on
the topic distributions.  We, instead, use the Fisher
kernel framework to expand the image representation
by decomposing the original BoW histogram into several
bags-of-words, one per topic, so that individual histogram
entries not only encode how often a word appears, but also
in combination with which other words it appears.  Whereas
compressed topic model representations were mostly found
to at best maintain BoW performance, we find significant
gains by using topic models. Finally, in contrast to the
PLSA Fisher kernel, which was previously studied as a
document similarity measure~\cite{chappelier09mlkdd,hofmann99nips}, we show that the LDA Fisher
kernel naturally involves discounting transformations.

Several other generative models have been proposed to
capture spatial regularities across image regions. For
example, the Spatial LDA model \cite{wang08nips} extends
the LDA model such that spatially neighboring visual
words are more likely assigned to the same topic.  The
counting grid model \cite{perina15pami}, which is a grid
of multinomial distributions, can be considered as an
alternative to the spatial topic models. In this approach,
the visual words of an image are treated as samples
from a latent local neighborhood of the counting grid.
Therefore, each local neighborhood of the model can be
interpreted as a spatial grid of topics. While these
studies show that incorporation of spatial information can
improve unsupervised semantic segmentation results
\cite{wang08nips}, or lead to better generative
classifiers compared to LDA \cite{perina15pami}, we limit
our focus to Fisher kernels of orderless, \ie exchangeable, generative
models in our study.

The computation of the LDA Fisher vector image
representation is technically more involved compared to
the PLSA model. In the case of the LDA model, the latent
model parameters cannot be integrated out analytically,
and the computation of the gradients is no longer
tractable. Similarly, the Fisher kernel for our Latent MoG
image model is intractable since the latent variables
(mixing weights, means, and variances) cannot be
integrated out analytically. We overcome this difficulty
by relying on the variational free-energy bound
\cite{jordan99ml}, which is obtained by subtracting the
Kullback-Leibler divergence between an approximate
posterior on the latent variables and the true posterior.
By imposing a certain independence structure on the
approximate posterior, tractable approximate inference
techniques can be devised. We then compute the gradient of
the variational bound as a surrogate for the intractable gradients of the 
exact log-likelihood.
The method of approximating Fisher kernels with the
gradient vector of a variational bound was first proposed
by Chandalia and Beal~\cite{chandalia06nipsw} in order to
obtain the LDA Fisher kernel. The only other work
incorporating this technique, to the best of our
knowledge, is the recent work of Perina
\etal~\cite{perina14icpr}, which proposes a variational
Fisher kernel for micro-array data. We show that
variational Fisher kernel is equivalent to the exact
Fisher vector when the variational bound is tight, and
demonstrate that in some cases it can be a mathematically more
convenient formulation, compared to the original Fisher
kernel definition.
Finally, we note that the variational approximation method
for Fisher kernels differs from Perina \etal
\cite{perina09nips}, which uses the variational
free-energy to define an alternative encoding, replacing
the Fisher kernel. 

In the following section we review the Fisher kernel
framework, and the variational approximation method. In
\sect{model} we present our non-\iid latent variable
models and propose novel Fisher vector representations
based on them. We present experimental results in
\sect{experiments}, and summarize our conclusions in
\sect{conclusion}.

%% file: fisher.tex
\section{Fisher vectors and variational approximation}
\label{sec:fisher}

In this section we present an overview of the
Fisher kernel framework, variational inference,
 and the variational Fisher kernel.

\subsection{Fisher vectors}

Images can be considered as samples from a generative
process, and therefore class-conditional generative models can be used for image categorization. However, it
is widely observed that discriminative classifiers 
 typically outperform
classification based on generative models, see \eg
\cite{halevy09is}. A simple explanation is that
discriminative classifiers aim to maximize the end goal,
which is to categorize entities based on their content.
In contrast, generative classifiers instead require
modeling class-conditional data distributions, which is
arguably a more difficult task than only learning  decision
surfaces, and therefore result in inferior categorization
performance.

The Fisher kernel framework proposed by Jaakkola and
Haussler \cite{jaakkola99nips} allows combining the power
of generative models and discriminative classifiers. In
particular, Fisher kernel provides a framework for
deriving a kernel from a probabilistic model. Suppose that
$p(\x)$ is a generative model with parameters
$\mathbf{\theta}$.\footnote{We drop the model parameters
    $\mathbf{\theta}$ from function arguments for
brevity.} Then, the Fisher kernel
$K(\x,\mathbf{x^\prime})$ is defined as 
\begin{equation}
K(\x,\mathbf{x^\prime})=g(\x)^\text{T}I^{-1}g(\mathbf{x^\prime})~,
\end{equation}
where the gradient $g(\x)=\nabla_{\theta}\log{p(\x)}$ is
called the {\em Fisher score}, and $I$ is the Fisher
information matrix:
\begin{equation}
I=\ex{\x\sim{p(\x)}}{g(\x)g(\x)^\text{T}}.
\end{equation}
 which is equivalent to the covariance of the Fisher score as computed using $p(\x)$, since $\ex{\x\sim{p(\x)}}{g(\x)}=\mathbf{0}$.
The inner product space (\ie explicit feature mapping)
induced by a Fisher kernel is given by
\begin{equation}
\phi(\x)=I^{-\frac{1}{2}} g(\x),
\end{equation}
where $I^{-\frac{1}{2}}$ is the whitening transform using
the Fisher information matrix.  S\'{a}nchez \etal
\cite{sanchez13ijcv}  refer to the normalized
gradient given by $\phi(\x)$ as the {\em Fisher vector}.
In practice the term ``Fisher vector'' is sometimes also used 
to refer to the non-normalized gradients (\ie Fisher score) as well.

The essential idea in Fisher kernel is to use gradients
$g(\x)$ of the data log-likelihood to extract features \wrt
a generative model. The Fisher information matrix, on the
other hand, is of lesser importance. A theoretical
motivation for using $I$ is that $I^{-1}g(\x)$ gives the
steepest descent direction along the manifold of the
parameter space, which is also known as the {\em natural
gradient}.  Another motivation is that $I$ makes the
Fisher kernel invariant to the re-parameterization
$\mathbf{\theta}\rightarrow{\psi(\mathbf{\theta})}$ for
any differentiable and invertible function $\psi$
\cite{bishop06patrec}, which can be easily shown using the
chain rule and the Jakobian matrix of the inverse function $\psi^{-1}$.

However, the computation of the Fisher information matrix
$I$ is intractable for many models. Although in principle it can be
approximated empirically as 
$I{\approx}\frac{1}{|X|}\sum_{\x\in{X}}{g(\x)g(\x)^\text{T}}$, 
the approximation itself can be costly 
if $g(\x)$ is high dimensional. In such cases, empirical
approximation can be used only for the diagonal terms.  
Alternatively, $I$ can be dropped altogether \cite{jaakkola99nips}
or analytical approximations can be derived, see \eg 
\cite{perronnin07cvpr,sanchez13ijcv,sanchez15prl}.

\subsection{Variational approximate inference}

Variational methods are a family of mathematical tools
that can be used to approximate intractable computations,
particularly those involving difficult integrals.
Originally developed in statistical physics based on the
{\em calculus of variations} and the {\em mean field
theory}, the variational approximation framework that we utilize
in this paper is known as the {\em variational 
inference}, and it is now among the most
successful approximate probabilistic
inference  techniques \cite{mackay03book,bishop06patrec,jordan99ml}.

In the context of probabilistic models, 
the central idea in variational methods is to 
devise a bound on the log-likelihood function 
in terms of an approximate 
posterior distribution over the latent variables.
Let $X$ denote the set of observed variables, and
$\Lambda$ denote the set of latent variables and latent parameters.
Suppose that $q(\Lambda)$ is an approximate distribution 
over the latent variables. Then, the distribution $p(X)$ can be 
decomposed as follows for any choice of the approximate posterior $q$:
\bea
\ln p(X) = F(p,q) + D\big( q || p\big). 
\label{eq:varmet:decompose}
\eea
In this equation, $F$ is the {\em variational free-energy} given by 
\bea 
F(p,q) & = & \int q(\Lambda) \ln \left(\frac{p(X,\Lambda)}{q(\Lambda)}\right) d\Lambda \\
& = & \ex{q(\Lambda)}{\ln p(X,\Lambda)} + H(q), 
\label{eq:varmet:F}
\eea
where $H(q)$ is the entropy of the distribution $q$. 
The term $D\left(q || p\right)$ in \Eq{varmet:decompose} is the Kullback-Leibler (KL) divergence
between the distributions $q(\Lambda)$ and $p(\Lambda|X)$:
\bea
D\big(q || p\big) = - \int q(\Lambda) \ln \left( \frac{p(\Lambda|X)}{q(\Lambda)} \right) d\Lambda .
\eea

Since the KL-divergence term $D\left(q || p\right)$ is strictly non-negative, the variational free energy  $F(p,q)$ is a
lower-bound on the true log-likelihood $\ln p(X)$, \ie $F(p,q) \leq \ln p(X)$. When the KL-divergence 
term is zero, \ie the distribution $q$ is equivalent to the true
posterior distribution, the bound $F$ is tight. 

In order to effectively utilize the decomposition in \Eq{varmet:F} for a given distribution $p$,
 we need to choose the distribution $q$ such that
it leads to a tractable and as tight as possible lower-bound $F(p,q)$.
For this purpose, we constrain $q$ to a family
of distributions $\mathcal Q$ that leads to tractable computations, typically by imposing  independence assumptions.
For example suppose that $\Lambda=(\lambda_1, \dots, \lambda_n)$, we may choose $\mathcal Q$ to be the set of distributions that factorize over the $\lambda_i$, \ie with $q(\Lambda)= \prod_{i=1}^n q_i(\lambda_i)$.
Given the family $\mathcal Q$, we maximize $F(p,q)$ by minimizing the KL divergence in \Eq{varmet:decompose} over all $q\in \mathcal Q$. 

\subsection{Variational Fisher kernel}
\label{sec:fisher:varfv}
 
In this paper, we utilize the variational free-energy 
bounds for two purposes. 
The first is to estimate the 
hyper-parameters of the LDA (\sect{LDA})
and the Latent MoG (\sect{LatMoG}) models from  training data using an approximate maximum likelihood procedure.
For this purpose, we iteratively update
the variational lower-bound with respect to the
variational distribution parameters, and the
model hyper-parameters; an approach 
that is known as the {\em variational expectation-maximization} procedure
\cite{jordan99ml}. 

Our second main use of the variational free-energy 
is to compute approximate Fisher vectors where the original Fisher vector is intractable to compute. In particular, 
we approximate the Fisher vector by the gradient of the 
variational lower-bound given by \Eq{varmet:F}, \ie
$g(\x) \approx \nabla_{\theta} F(p,q)$, which we refer 
to as {\em variational Fisher vector}. Since, the entropy
$H(q)$ is constant \wrt model parameters, the variational
Fisher vector $\theta_q$ can equivalently be written as
\bea
\phi_q(X) = I^{-\frac{1}{2}} \nabla_{\theta} \ex{q}{\ln p(X,\Lambda)}.
\eea
where $I$ is the (approximate) Fisher information matrix.

We have already discussed that the variational bound in \Eq{varmet:F} is tight when the distribution $q$ 
matches the posterior on the hyper-parameters. We will now show that its
gradient equals that of the data log-likelihood if the bound is tight. In order to prove this, we first write the partial derivative of the lower-bound with
respect to some model (hyper-)parameter $\theta$:
\bea
\grad{F}{\theta} & = &  \grad{\ex{q}{\ln p(X,\Lambda)}}{\theta}.
\eea
By definition,
we can interchange the differential operator and the expectation:
\bea
\grad{F}{\theta}  & = &  \ex{q}{\grad{ \ln p(X,\Lambda)}{\theta} }.
\eea
Without loss of generality, we assume that all latent variables are  continuous, in which case the expectation is equivalent to
\bea
\grad{F}{\theta} & = &  \int{q(\Lambda) \grad{\ln p(X,\Lambda)}{\theta}}d\Lambda.
\eea
By following differentiation rules, we obtain the 
 equation:
\bea
\grad{F}{\theta} & = &  \int{q(\Lambda) \frac{1}{p(\Lambda|X)p(X)} \grad{p(X,\Lambda)}{\theta}}d\Lambda.
\eea

Since the bound is assumed to be tight, the $q(\Lambda)$ and $p(\Lambda|X)$ are identical. In addition, we observe that $p(X)$ is a constant with respect to the integration variables. Therefore, we can simplify the equation as follows:
\bea
\grad{F}{\theta} & = &  \frac{1}{p(X)}{\int \grad{p(X,\Lambda)}{\theta} d\Lambda}, 
\eea
which can be re-written as follows:
\bea
\grad{F}{\theta} & = &  \frac{1}{p(X)}\grad{\int p(X,\Lambda)d\Lambda}{\theta}. 
\eea
Finally, we integrate out $\Lambda$ and simplify the 
equation into the following form:
\bea
\grad{F}{\theta} & = &  \grad{\ln p(X)}{\theta},
\eea
which completes the proof.

In addition to presenting a relationship between the
original Fisher vector and the variational Fisher vector
definitions, the proof shows that the latter formulation 
can be used as an alternative framework. In fact,
we observe that the variational formulation can in some
cases be mathematically more convenient to derive 
Fisher vector representations.
Even though our main interest in this paper is to compute
approximate representations based on the LDA and latent MoG image models presented in the next section,  we present two additional
examples in \app{varfv} that demonstrate the usefulness of
the variational formulation.

%% file: model.tex
\section{Non-iid image representations}
\label{sec:model}

In this section we present our non-\iid models for local image descriptors.  We start
with a model for BoW quantization indices, and 
 extend the model to capture co-occurrence
statistics across visual words using LDA in \sect{LDA}. 
Finally, we consider a non-\iid extension of mixture of Gaussian models over sets of local descriptors in \sect{LatMoG}.

\input{model_word}

\input{model_topic}

\input{model_gauss}

%% file: model_word.tex
\subsection{Bag-of-words and the multivariate P\'{o}lya model}
\label{sec:polya}

The standard BoW image representation can be interpreted
as applying the Fisher kernel framework to a simple \iid
multinomial model over visual word
indices, as shown in~\cite{krapac11iccv}.  Let $w_{1:N} = \{w_1, \dots
,w_N\}$  denote the visual word indices corresponding to
$N$  patches sampled in an image, and  let $\pi$ be a
learned multinomial over $K$ visual words, parameterized
in log-space, \ie $p(w_i=k)=\pi_k$ with
$\pi_k=\exp(\gamma_k) / \sum_{k'} \exp(\gamma_{k'})$. 
The data likelihood for the BoW model is given by 
\begin{equation}
    p(w_{1:N}) = \prod_{i=1}^N p(w_i=k).
\label{eq:bow}
\end{equation}
The 
gradient of the data log-likelihood is in this case given
by 
\begin{equation}
    \grad{\sum_{i=1}^N \ln p(w_i)}{\gamma_k} = n_k -N\pi_k,
\label{eq:llbow}
\end{equation}
where  $n_k$ denotes the number of occurrences of visual
word $k$ among  the set of indices $w_{1:N}$.  This is a
shifted version of the standard BoW histogram, where the
mean of all image representations is centered at the
origin.  We stress that this multinomial interpretation of
the  BoW  model assumes that the visual word indices
across all images are \iid, which directly generates the product form in the likelihood of \Eq{bow}, and the count statistic in the gradient of the log-likelihood in \Eq{llbow}.

\def\mygraphmodel#1{\includegraphics[scale=.77]{model/#1}}
\def\mypicwithcaplabel#1#2#3{%
    \subfloat[#2]{%
    \mygraphmodel{#1}
    \label{#3}
    }}

\begin{figure}
\begin{center}
\subfloat[Multinomial BoW model]{%
    \begin{minipage}[b]{4cm}
    \begin{center}
    \mygraphmodel{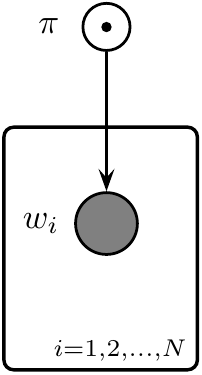}
    \end{center}
    \end{minipage}
    \label{fig:GraphModWord:BoW}
}
\hspace{0.5cm}
\mypicwithcaplabel{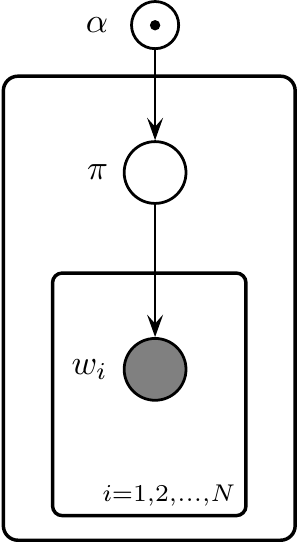}{P\'olya model}{fig:GraphModWord:LaBoW}
\end{center}
\caption{Graphical representation of the models in
\sect{polya}: (a)
multinomial BoW model, (b) P\'olya model.
The outer plate in (b) refer
to images. The index $i$ runs over the visual word indices
in an image.  Nodes of
observed variables are shaded, and those of
(hyper-)parameters are marked with a central dot in the
node.}
\label{fig:GraphModWord}
\end{figure}

Our first non-\iid model assumes that for each image there is a different, a-priori
unknown, multinomial generating the visual word indices in 
that image. 
In this model visual word indices within an image are mutually dependent, 
since knowing some of the $w_i$ provides information on the underlying multinomial $\pi$,
 and thus also provides information on which subsequent indices  could be sampled from it. 
The model is parameterized by a non-symmetric Dirichlet prior over the latent image-specific multinomial,  $p(\pi)=\mathcal{D}(\pi|\alpha)$ with $\alpha= (\alpha_1,\dots, \alpha_K)$, and the $w_i$ are modeled as \iid samples from $\pi$.
The marginal distribution on the $w_i$ is obtained by integrating out $\pi$:
\bea
p(w_{1:N}) & = &  \int p(\pi) \prod_{i=1}^N p(w_i|\pi) d\pi.
\eea
This model is known as the multivariate P\'{o}lya, or Dirichlet
compound multinomial \cite{madsen05icml}, and the integral simplifies to 
\bea
p(w_{1:N}) & = & \frac{\Gamma(\hat{\alpha})}{\Gamma(N+\hat{\alpha})}\prod_{k=1}^K\frac{\Gamma(n_k+\alpha_k)}{\Gamma(\alpha_k)},
\eea
where $\Gamma(\cdot)$ is the Gamma function, and $\hat{\alpha}=\sum_{k=1}^K \alpha_k$.
See \fig{GraphModWord:BoW} and
\fig{GraphModWord:LaBoW} for a graphical
representation of the BoW multinomial model, and the
P\'olya model.

Following the Fisher kernel framework, we represent an image by the gradient \wrt the hyper-parameter $\alpha$ of the log-likelihood of the visual word indices $w_{1:N}$.
The partial derivative \wrt $\alpha_k$ is given by 
\bea
\hspace{-1mm}
\grad{\ln p(w_{1:N})}{\alpha_k}\!=\!\psi(\alpha_k\!+\!n_k)\!-\!\psi(\hat{\alpha}\!+\!N)\!-\!\psi(\alpha_k)\!+\!\psi(\hat{\alpha}), \label{eq:GradLaBow}
\eea
where $\psi(x) = \partial \ln \Gamma(x) / \partial x$ is the digamma function.

Only the first two terms in \Eq{GradLaBow}  depend on the counts $n_k$, and for fixed $N$ the gradient is determined up to additive constants by $\psi(\alpha_k+n_k)$, \ie it is given by a transformation of the  visual word counts $n_k$.
\fig{transfer} shows the transformation $\psi(\alpha+n)$ for various values of $\alpha$, along with the square-root function used in the Hellinger distance for reference.
We see that  the same monotone-concave discounting effect is obtained as by  taking the square-root of histogram entries. 
This transformation arises naturally in our latent variable model,
and suggests that such transformations are successful \emph{because} they correspond 
to a more realistic non-\iid model, \cf \fig{iid}.

Observe that in the limit of $\alpha\rightarrow \infty$ the transfer function becomes linear,
since for large $\alpha$ the Dirichlet prior tends to a delta peak on the multinomial simplex and thus removes the uncertainty on the underlying multinomial, with an  observed multinomial BoW model as its limit.
In the limit of $\alpha\rightarrow 0$, corresponding to priors that concentrate their mass at sparse multinomials,  the transfer function becomes a step function. 
This is intuitive, since in the limit of ultimately sparse distributions only one word will be observed, and  its count    no longer matters, we only need to know which word is observed to determine which $\alpha_k$ should be increased to improve the log-likelihood.

\def\mysingleplot#1{\includegraphics[width=.6\linewidth]{#1}}
\begin{figure}
\begin{center}
\mysingleplot{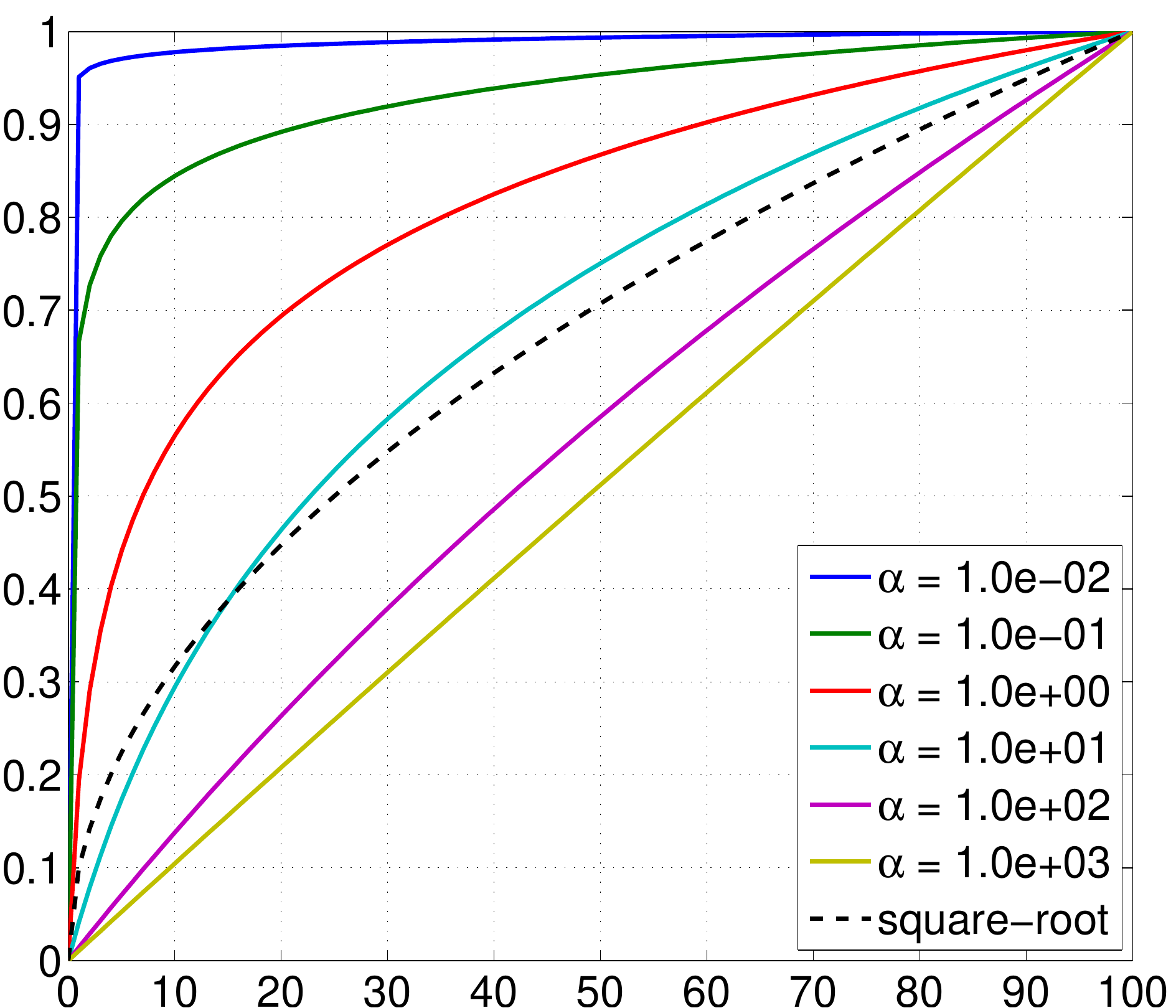}
\end{center}
\caption{Digamma functions $\psi(\alpha+n)$ for various $\alpha$, and   $\sqrt n$  as a function of $n$;
 functions  have been rescaled to the range $[0,1]$.}
\label{fig:transfer}
\end{figure}

%% file: model_topic.tex
\subsection{Capturing  co-occurrence with topic models}
\label{sec:LDA}

\begin{figure}
\hspace{1.2cm}
\mygraphmodel{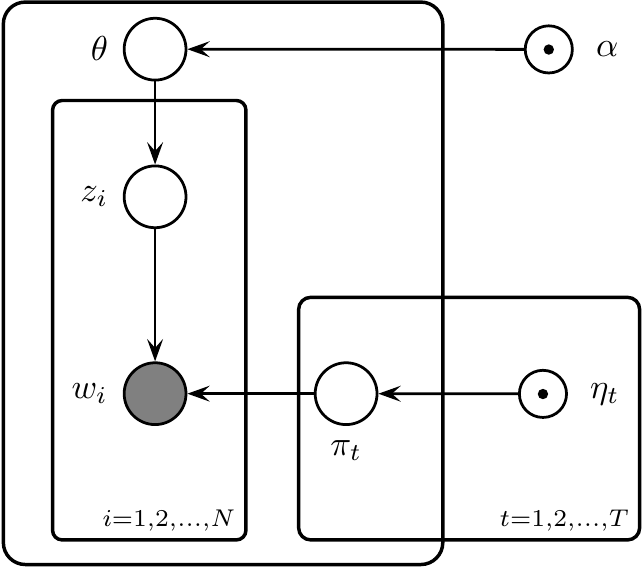}
\caption{Graphical representation of LDA. The outer plate refers to images. The index $i$ runs over patches, and index $t$ over topics.}
\label{fig:LDA}
\end{figure}

The P\'olya model is non-\iid but it does not model co-occurrence across visual words, this can be seen from the posterior distribution $p(w=k|w_{1:N}) = \int p(w=k|\pi)p(\pi|w_{1:N})d\pi\propto n_k+\alpha_k $.
The model just predicts to see more visual words of the type it has already seen before.
In our second model, we extend the P\'olya model to capture co-occurrence statistics of visual words using latent Dirichlet allocation (LDA) \cite{blei03jmlr}.
We model the visual words in an image as a mixture of $T$ topics, encoded by a multinomial  $\theta$ mixing the  topics, where each topic itself is represented by a multinomial distribution $\pi_t$ over the $K$ visual words.
We associate a variable $z_i$, drawn from $\theta$, with each patch that indicates which topic  was used to draw its visual word index $w_i$.
We place Dirichlet priors on the topic mixing, $p(\theta)  =  \mathcal{D}(\theta | \alpha)$,  and the topic distributions $p(\pi_t)=\mathcal{D}(\pi_t | \eta_t)$,
and  integrate these out to obtain the marginal distribution over  visual word indices as:
\bea
p(w_{1:N}) &= & \iint p(\theta) p(\pi) \prod_{i=1}^N p(w_i|\theta,\pi) d\theta d\pi,\\
p(w_i=k|\theta,\pi) & = & \sum_{t=1}^T p(z_i=t|\theta)p(w_i=k|\pi_t).
\eea
See \fig{LDA} for a graphical representation of the model. 
Note that this model is equivalent to the P\'olya model discussed above when there is only a single topic, \ie for $T=1$.

Both  the log-likelihood and its gradient are  intractable to compute for the LDA model. 
As discussed in \sect{fisher:varfv}, however, we can resort to variational methods to compute a free-energy bound $F$ using an approximate posterior.
Here we use a completely factorized approximate posterior as in~\cite{blei03jmlr} of the form 
\bea
q(\theta,\pi_{1:T},z_{1:N}) = q(\theta)  \prod_{t=1}^T q(\pi_t)   \prod_{i=1}^N q(z_i).
\eea
The update equations of the variational distributions $q(\theta)=\mathcal{D}(\theta|\alpha^*)$ and $q(\pi_t) = \mathcal{D}(\pi_t|\eta_t^*)$ to maximize the free-energy bound $F$ are given by:
\bea
\alpha_t^*  = \alpha_t  + \sum_{i=1}^N q_{it},\quad\quad\quad
\eta_{tk}^*  =  \eta_{tk} + \sum_{i:w_i=k} q_{it},
\eea
where $q_{it} = q(z_i=t)$, which is itself updated according to $q_{it} \propto \exp[ \psi(\alpha^*_t) - \psi(\hat{\alpha}^*) + \psi(\eta_{tk}^*) -  \psi(\hat{\eta}_{t}^*)  ].$
These update equations can be applied iteratively to monotonically improve the variational bound.

The gradients of $F$ \wrt the hyper-parameters are obtained from these as
\bea
\grad{F}{\alpha_t} & = & \psi(\alpha_t^*) - \psi(\hat{\alpha}^*) - [ \psi(\alpha_t) - \psi(\hat{\alpha})],\\
\grad{F}{\eta_{tk}} & = & \psi(\eta_{tk}^*) - \psi(\hat{\eta}_{t}^*) - [ \psi(\eta_{tk}) - \psi(\hat{\eta}_{t})].
\eea
The gradient \wrt $\alpha$ encodes a discounted version of the topic proportions as they are inferred in the image.
The gradients \wrt the hyper-parameters $\eta_t$ can be interpreted as decomposing the bag-of-word histogram  over the $T$ topics, and encoding the soft counts of words assigned to each topic. 
The entries $\grad{F}{\eta_{tk}}$ in this representation not only code how often a word was observed but also in combination with which other words, since the co-occurrence of words throughout the  image will determine the inferred topic mixing and thus the word-to-topic posteriors $q_{it}$.

In our experiments we  compare  LDA with the PLSA model \cite{hofmann01ml}. 
This model treats the topics $\pi_t$, and the topic mixing $\theta$ as non-latent parameters which are estimated by maximum likelihood. 
To represent images using PLSA we apply the Fisher kernel framework and compute gradients of the log-likelihood \wrt $\theta$ and the $\pi_t$. 
The PLSA model with a single topic reduces to the \iid multinomial model discussed in the previous section.

%% file: model_gauss.tex

\subsection{Modeling descriptors using latent MoG  models}
\label{sec:LatMoG}

In this section we turn to the image representation of Perronnin and Dance \cite{perronnin07cvpr} that applies the Fisher kernel framework to  mixture of Gaussian (MoG) models over  local descriptors. An improved version of this representation using power normalization was presented in \cite{perronnin10eccv}.  

\begin{figure}
\mypicwithcaplabel{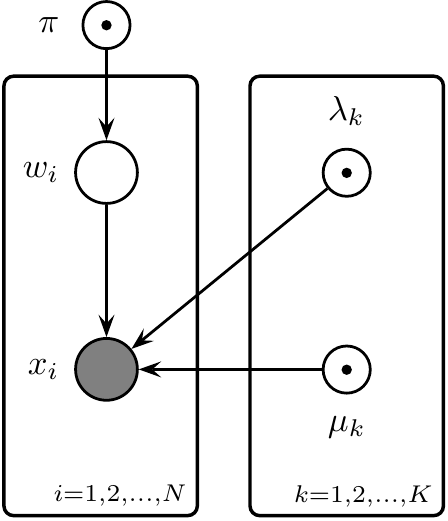}{MoG model}{fig:GraphModGauss:MoG}
\hfill
\mypicwithcaplabel{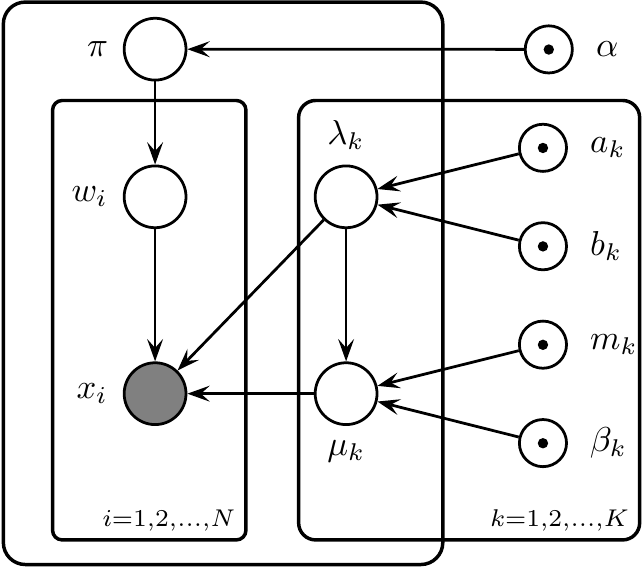}{Latent MoG model}{fig:GraphModGauss:LatMoG}
\caption{Graphical representation of the models in
\sect{LatMoG}: (a) MoG model,
(b) latent MoG model. The outer plate in (b) without indexing
refer to images. The index $i$ runs over the local descriptors, and index $k$ over Gaussians in the mixture which represent the visual words.  
}
\label{fig:GraphModGauss}
\end{figure}

A MoG density $p(x) = \sum_{k=1}^K \pi_k
\mathcal{N}(x;\mu_k,\sigma_k)$ is defined by mixing
weights $\pi=\{\pi_k\}$, means $\mu=\{\mu_k\}$ and
variances $\sigma=\{\sigma_k\}$.\footnote{We present here
the uni-variate case for clarity, extension to the
multivariate case with diagonal covariance matrices is
straightforward.} The $K$ Gaussian components of the
mixture correspond to the $K$ visual words in a BoW model.
In \cite{perronnin07cvpr,perronnin10eccv}, local descriptors across images
are assumed to be \iid samples from a single MoG model
underlying all images. 
They represent an image by the
gradient of the log-likelihood of the extracted local descriptors
$x_{1:N}$ \wrt the model parameters. Using the soft-assignments $p(k|x) = \pi_k \mathcal{N}(x;\mu_k,\sigma_k) / p(x)$ of local descriptors to mixture components the partial derivatives are computed as:
\begin{align}
\grad{\ln p(x_{1:N})}{\gamma_k}&=\sum_{i=1}^N{p(k|x_i)-\pi_k},\\
\grad{\ln p(x_{1:N})}{\mu_k}&=\sum_{i=1}^Np(k|x_i)(x-\mu_k) / \sigma_k,\\
\grad{\ln p(x_{1:N})}{\lambda_k}&=\sum_{i=1}^N p(k|x_i)\left(\sigma_k-(x_i-\mu_k)^2\right)/2,
\end{align}
where we re-parameterize the mixing weights
as $\pi_k=\exp(\gamma_k) / \sum_{k'=1}^K \exp(\gamma_{k'})$,
and the Gaussians with precisions
$\lambda_k=\sigma_k^{-1}$, as in \cite{krapac11iccv}.  For
local descriptors of dimension $D$,
the gradient yields an image representation of size
$K(1+2D)$, since for each of the $K$ visual words there is
one derivative \wrt its mixing weight, and $2D$
derivatives for the means and variances in the $D$
dimensions.  This representation thus stores more
information about the local descriptors assigned to a visual
word than just their count, as a result higher recognition performance
can be obtained using the same number of visual words as compared to the BoW representation.

In analogy to the P\'{o}lya model, we remove the \iid assumption by defining a MoG model per image and treating its parameters as latent variables.
We place  conjugate priors on the image-specific parameters: a Dirichlet prior on the mixing weights, $p(\pi)  =  \mathcal{D}(\pi | \alpha)$, and a combined Normal-Gamma prior on the means $\mu_k$ and precisions $\lambda_k=\sigma^{-1}_k$:
\bea
p(\lambda_k)       & = & \mathcal{G}(\lambda_k|a_k,b_k),\\
p(\mu_k|\lambda_k) & = & \mathcal{N}(\mu_k|m_k,(\beta_k\lambda_k)^{-1}).
\eea
The distribution on the descriptors $x_{1:N}$ in an image is obtained by integrating out
the latent MoG parameters:
\begin{align}
& p(x_{1:N})  =  \iiint p(\pi)p(\mu,\lambda) \prod_{i=1}^N p(x_i|\pi,\mu,\lambda) d\pi d\mu d\lambda,\\
& p(x_i|\pi,\mu,\lambda) = \sum_{k=1}^K p(w_i\!=k|\pi)p(x_i|w_i\!=k,\lambda,\mu),
\end{align}
where $p(w_i\!=\!k|\pi)\!=\!\pi_k$, and $p(x_i|w_i\!=\!k,\lambda,\mu)\!=\!\mathcal{N}(x_i|\mu_k,\lambda_k^{-1})$ is the Gaussian  corresponding to the $k$-th visual word.
See \fig{GraphModGauss:MoG} and
\fig{GraphModGauss:LatMoG}  for  graphical
representations of the MoG model and the latent MoG model.

Computing the log-likelihood in this model
is also intractable, as is computing the gradient
of the log-likelihood which we need for both
hyper-parameter learning and to extract the Fisher vector
representation.  
To overcome these problems we replace the intractable log-likelihood with its variational
lower bound.

By constraining the variational posterior $q$ in the bound $F$ given by \Eq{varmet:F} to factorize as $q(\pi,\mu,\lambda,w_{1:N}) =  q(\pi,\mu,\lambda)q(w_{1:N}) $ over the latent MoG parameters and the assignments of local descriptors to visual words, we obtain a bound for which we can tractably compute its value and gradient \wrt the hyper-parameters. 
Given this factorization it is easy to show that the optimal $q$ will further factorize as
\bea
q(\pi,\mu,\lambda,w_{1:N}) =  q(\pi)\prod_{k=1}^K q(\mu_k|\lambda_k)q(\lambda_k)\prod_{i=1}^N q(w_i),
\eea
and that the variational posteriors on the model parameters will have the form of Dirichlet and Normal-Gamma distributions
 $q(\pi)  =  \mathcal{D}(\pi|\alpha^*)$,  $q(\lambda_k) =  \mathcal{G}(\lambda_k| a^*_k, b^*_k)$, $q(\mu_k|\lambda_k)  =  \mathcal{N}(\mu_k| m_k^*, (\beta_k^*\lambda_k)^{-1})$.
Given the hyper-parameters we can update the variational distributions to maximize the variational lower  bound.
In order to write the update equations, it is convenient to define the following  sufficient statistics :
\bea
s_k^0 = \sum_{i=1}^N q_{ik},\quad s_k^1 = \sum_{i=1}^N q_{ik}x_i, \quad s_k^2 = \sum_{i=1}^N q_{ik}x_i^2.
\label{eq:latmog:sufstat}
\eea
where $q_{ik} = q(w_i=k)$. Then, the parameters of the
optimal variational distributions on the MoG parameters
for a given image are found as:
\bea
%
%
\alpha^*_k & = & \alpha_k + s^0_k, \label{eq:up_alpha} \\
%
%
\beta_k^* & = & \beta_k+s_k^0, \label{eq:up_beta}\\
m_k^* &= & (s_k^1 + \beta_k m_k) / \beta_k^*, \label{eq:up_mean}\\
%
%
a^*_k & = & a_k+s^0_k/2, \label{eq:up_a}\\
b^*_k & = & b_k+\half(\beta_k m_k^2 + s_k^2) - \half\beta^*_k(m_k^*)^2.
\eea
The component assignments $q(w_i)$ that maximize the bound given the variational distributions on the MoG parameters are given by:
\bea
\ln q_{ik} & = & \ex{q(\pi)q(\lambda_k,\mu_k)}{\ln \pi_k + \ln \mathcal{N}(x_i|\mu_k,\lambda_k^{-1})}\\
	   & = & \psi(\alpha_k^*) - \psi(\hat{\alpha}^*) +\half\Big[ \psi(a_k^*)-\ln b_k^*   \Big] \\
& & -\half\Big[  \frac{a_k^*}{b_k^*} (x_i-m^*_{k})^2 +  (\beta_k^*)^{-1}   \Big].
\eea
Since the sufficient statistics given by \Eq{latmog:sufstat} depend on the component assignments, the distributions on the MoG parameters and the component assignments can be updated iteratively to improve the bound.

Using the above variational update equations, we obtain the variational distribution, and therefore the lower-bound on the log-likelihood for each image. During training, we learn the model hyper-parameters by iteratively maximizing the sum of the lower-bounds for the training images \wrt the hyper-parameters, and \wrt the variational parameters. 
Once the latent MoG model is trained, we use the per-image lower-bounds to extract the approximate Fisher vector descriptors according to the gradient of $F$ with respect to the model hyper-parameters.

The gradient of $F$ \wrt the hyper-parameters  depends only on the variational distributions on the MoG parameters of an image $q(\pi)$,  $q(\lambda_k) $, and $q(\mu_k|\lambda_k)$, and not on the component assignments $q(w_i)$.
For the precision hyper-parameters we find:
\bea
\grad{F}{a_k} &= & \left[\psi(a_k^*)-\ln b_k^*\right] - \left[\psi(a_k) -\ln b_k\right], \\
\grad{F}{b_k} &= & \frac{a_k}{b_k}-\frac{a^*_k}{b^*_k},
\eea
For the hyper-parameters of the means:
\bea
\grad{F}{\beta_k} &\!=\!& \half\left(\beta^{-1}_k -  \frac{a_k^*}{b_k^*}(m_k-m_k^*)^2 - 1/\beta_k^* \right),\\
\grad{F}{m_k} & = & \beta_k \frac{a_k^*}{b_k^*}(m_k^*-m_k),\label{eq:gradmean}
\eea
and for the hyper-parameters of the mixing weights:
\bea
\grad{F}{\alpha_k} & = & \left[\psi(\alpha^*_k) - \psi(\hat{\alpha}^*)\right] -\left[\psi(\alpha_k) - \psi(\hat{\alpha})\right].\label{eq:gradpi}
\eea

By substituting the update equation \eq{up_alpha} for the variational parameters $\alpha_k^*$ in the gradient   
\Eq{gradpi}, we exactly recover the gradient of the multivariate P\'{o}lya model, albeit using soft-counts $s_k^0=\sum_{i=1}^N q(w_i\!=\!k)$ of visual word occurrences here. 
Thus, the bound leaves  the qualitative behavior of the multivariate P\'{o}lya model intact.
Similar discounting effects can be observed in the gradients of the hyper-parameters of the means and variances.
Substitution of the update equation \eq{up_mean} for the variational parameters $m_k^*$ in the gradient  \Eq{gradmean}, reveals that the gradient is similar to the square-root of the gradient obtained in \cite{perronnin07cvpr} for the MoG mean parameters. The  discounting function for this gradient is however slightly different from the  $\psi(\cdot)$ function, but has a similar monotone concave form. We consider examples of the learned discounting  functions in \sect{xp:mog}. 

Our latent MoG model associates two hyper-parameters $(m_k,\beta_k)$ with each mean $\mu_k$, and similar for the precisions. 
Therefore, our image representation are almost twice as long compared to the \iid MoG model: $K(1+4D)$ \vs  $K(1+2D)$ dimensions. 
The updates of the variational parameters $\beta_k^*$ and $a_k^*$ in equations (\ref{eq:up_beta}) and (\ref{eq:up_a}), however, only involve the zero-order statistics $s_k^0$. 
In \cite{perronnin07cvpr} the FV components corresponding to the mixing weights of the MoG, which are also based on zero-order statistics, were shown to be redundant when also including the components corresponding to the means and variances.
Therefore, we expect the gradients \wrt the corresponding hyper-parameters $\beta_k$ and $a_k$ to be of little importance for image classification purposes. 
Experimental results, not reported here, have empirically verified this. 
We therefore fix the number of Gaussians rather than the FV dimension when we compare different representations in the next section, and use all FV components to avoid confusion.

%% file: experiments.tex
\section{Experimental evaluation}
\label{sec:experiments}

In this section, we present a detailed evaluation of the
latent BoW, LDA and the latent MoG models over SIFT local
descriptors using the PASCAL VOC'07~\cite{everingham15ijcv} data set
in \sect{xp:bow}, \sect{xp:topic} and \sect{xp:mog},
respectively.  Then, we present a
empirical study on the relationship between the model
likelihood and image categorization performance in \sect{xp:lhood}. Finally,
we evaluate the Latent MoG model,
which is the most advanced model that we consider, over the CNN-based local
descriptors, and compare against the state-of-the-art on
the PASCAL VOC'07 and MIT Indoor~\cite{quattoni09cvpr}
data sets in \sect{xp:cnn}.

Now, we first describe our experimental setup for the
SIFT-based experiments used in the subsequent sections.

\subsection{Experimental setup}
\label{sec:xp:setup}

In order to extract SIFT descriptors, we use the
experimental setup described in the evaluation paper of
Chatfield \etal \cite{chatfield11bmvc}: we sample local
SIFT descriptors from the same dense grid (3 pixel stride,
across 4 scales), which results in around $60,000$ patches
per image, project the local descriptors to $80$
dimensions with PCA, and train the MoG visual vocabularies
from $1.5\times10^6$ descriptors. For the PASCAL VOC'07
data set, we use the interpolated mAP score specified by
the VOC evaluation protocol~\cite{everingham15ijcv}. 

We compare global image representations, and representations
that capture spatial layout by concatenating the
signatures computed over various spatial cells as in the
spatial pyramid matching (SPM) method
\cite{lazebnik06cvpr}.  Again, we follow
\cite{chatfield11bmvc} and combine a $1\times 1$, a
$2\times 2$, and a $3 \times 1$ grid.  Throughout, we use
linear SVM classifiers, and we cross-validate the
regularization parameter.

Before training the classifiers we apply two
normalizations to the image representations. First, we
whiten the representations so that each dimension is
zero-mean and has unit-variance across images in order to
approximate normalization with the inverse Fisher
information matrix. Second, following
\cite{perronnin10eccv}, we also $\ell_2$ normalize the
image representations.

For the BoW, PLSA and MoG models, we compare using Fisher vectors with and without power normalization, and to using the Fisher vectors of the corresponding latent variable models.  As in \cite{perronnin10eccv}, power
normalization is applied after whitening, and before $\ell_2$
normalization.
We evaluate two types of power normalization: (i) signed square-rooting
($\rho=1/2$) as in \cite{perronnin10eccv,chatfield11bmvc}, which we denote by
a prefix ``Sqrt'', (ii) more general power normalization, which we denote by
a prefix ``Pn''. In the latter case, we cross-validate the parameter $\rho \in \{0,0.1,0.2,...,1\}$ for each setting, but keeping it fixed across the classes.

In Tables \ref{tab:VOC07_bowvslbow_overK}, \ref{tab:VOC07_gmmvslgmm_overK},
\ref{tab:VOC07_cnn} and \ref{tab:MITIndoor_cnn}, the bold numbers
indicate the top performing representations in each setting that
are statistically equivalent, which we measure by using 
the bootstrapping method proposed in Everingham \etal~\cite{everingham15ijcv}, at $95\%$ confidence interval. 
In Tables \ref{tab:VOC07_sota} and \ref{tab:MITIndoor_sota}, we are unable to run the test on other state-of-the-art approaches, as the statistical significance test requires original classification scores on the test images.

\input{experiments_bow}

\input{experiments_topic}

\input{experiments_mog}

\input{experiments_lhoodVsMAP}

\input{experiments_cnn}

%% file: experiments_bow.tex
\subsection{Evaluating BoW and P\'olya models}
\label{sec:xp:bow}

\input{tables/voc07_bow}

\begin{figure}
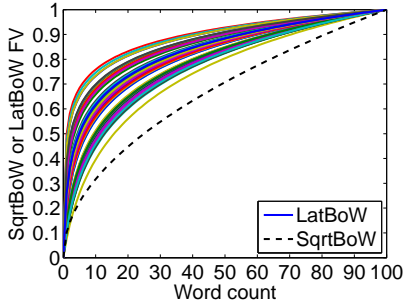

\begin{center}
\mysingleplot{learnt_curves/polya_synthetic}
\end{center}
\caption{Comparison of the discounting functions learned by the latent BoW model for 64 visual words (solid), and the square-root transformation (dashed).
Transformed counts are rescaled to the range $[0,1]$.}
\label{fig:PolyaExampleCurves}
\end{figure}

In \tab{VOC07_bowvslbow_overK} we compare the results
obtained using standard BoW histograms, two types of power normalized
histograms, and the P\'{o}lya model. 
In all three cases, we generate the visual word counts from soft assignments of patches to  the MoG components. 
Overall, we see that the spatial information
of SPM is useful, and that larger vocabularies increase
performance. We observe that both power normalization and the
P\'olya model both consistently improve the BoW
representation, across all dictionary sizes, and with or
without SPM. Furthermore, the P\'olya model generally
leads to larger improvements than power normalization. These
results are in line with the observation of \sect{polya} that the
non-\iid P\'olya model generates similar transformations
on BoW histograms as power normalization does, and show that normalization by the digamma function is at least as effective as power normalization.

\fig{PolyaExampleCurves} illustrates the discounting
functions  learned by the P\'{o}lya model for a
 dictionary of  $64$ visual words, without a spatial pyramid.
Each solid curve in the figure corresponds to one of the
visual words, and shows the corresponding digamma function 
$\psi(\alpha_k+n_k)$ as a function of the
visual word count $n_k$.  Compared to the square-root
transformation, which is shown by the dashed curve, we
observe that the P\'{o}lya model generally leads to similar but somewhat stronger
discounting effect.

%% file: tables/voc07_bow.tex
\begin{table}
\begin{center}
{\footnotesize
\input{autogen/mkAllFiguresAndTables/voc07_bow.tex}
}
\end{center}
\caption{
Comparison of representations with and without SPM:  BoW, two types of power normalized BoW, and P\'olya. 
} 
\label{tab:VOC07_bowvslbow_overK}
\end{table}

%% file: autogen/mkAllFiguresAndTables/voc07_bow.tex
\begin{tabular}{ccccccc}
\hline
\multicolumn{1}{|c|}{SPM} & \multicolumn{1}{c||}{Method}  & \multicolumn{1}{c|}{64} & \multicolumn{1}{c|}{128} & \multicolumn{1}{c|}{256} & \multicolumn{1}{c|}{512} & \multicolumn{1}{c|}{1024} \\
\hline
\noalign{\smallskip}
\hline
\multicolumn{1}{|c|}{No} & \multicolumn{1}{c||}{BoW} & \multicolumn{1}{c|}{21.0} & \multicolumn{1}{c|}{28.6} & \multicolumn{1}{c|}{37.1} & \multicolumn{1}{c|}{40.5} & \multicolumn{1}{c|}{43.7}\\
\hline
\multicolumn{1}{|c|}{No} & \multicolumn{1}{c||}{SqrtBoW} & \multicolumn{1}{c|}{20.8} & \multicolumn{1}{c|}{28.4} & \multicolumn{1}{c|}{\textbf{37.6}} & \multicolumn{1}{c|}{\textbf{41.4}} & \multicolumn{1}{c|}{\textbf{46.0}}\\
\hline
\multicolumn{1}{|c|}{No} & \multicolumn{1}{c||}{PnBoW} & \multicolumn{1}{c|}{20.9} & \multicolumn{1}{c|}{\textbf{30.4}} & \multicolumn{1}{c|}{37.4} & \multicolumn{1}{c|}{\textbf{41.5}} & \multicolumn{1}{c|}{\textbf{46.3}}\\
\hline
\multicolumn{1}{|c|}{No} & \multicolumn{1}{c||}{LatBoW} & \multicolumn{1}{c|}{\textbf{21.7}} & \multicolumn{1}{c|}{\textbf{30.0}} & \multicolumn{1}{c|}{\textbf{38.4}} & \multicolumn{1}{c|}{\textbf{41.0}} & \multicolumn{1}{c|}{44.9}\\
\hline
\noalign{\smallskip}
\hline
\multicolumn{1}{|c|}{Yes} & \multicolumn{1}{c||}{BoW} & \multicolumn{1}{c|}{37.1} & \multicolumn{1}{c|}{39.8} & \multicolumn{1}{c|}{42.8} & \multicolumn{1}{c|}{46.3} & \multicolumn{1}{c|}{48.9}\\
\hline
\multicolumn{1}{|c|}{Yes} & \multicolumn{1}{c||}{SqrtBoW} & \multicolumn{1}{c|}{37.9} & \multicolumn{1}{c|}{\textbf{41.3}} & \multicolumn{1}{c|}{\textbf{44.6}} & \multicolumn{1}{c|}{47.8} & \multicolumn{1}{c|}{\textbf{51.6}}\\
\hline
\multicolumn{1}{|c|}{Yes} & \multicolumn{1}{c||}{PnBoW} & \multicolumn{1}{c|}{37.7} & \multicolumn{1}{c|}{\textbf{41.4}} & \multicolumn{1}{c|}{\textbf{44.6}} & \multicolumn{1}{c|}{47.4} & \multicolumn{1}{c|}{51.3}\\
\hline
\multicolumn{1}{|c|}{Yes} & \multicolumn{1}{c||}{LatBoW} & \multicolumn{1}{c|}{\textbf{39.5}} & \multicolumn{1}{c|}{\textbf{41.8}} & \multicolumn{1}{c|}{\textbf{45.4}} & \multicolumn{1}{c|}{\textbf{49.2}} & \multicolumn{1}{c|}{\textbf{52.3}}\\
\hline
\end{tabular}

%% file: experiments_topic.tex
\subsection{Evaluating  topic model representations} 
\label{sec:xp:topic}

We compare different topic model representations of \sect{LDA}:
 Fisher vectors computed on the PLSA model, its power normalized version, 
and using the corresponding LDA latent variable model. We compare to the corresponding BoW representations, and include SPM in all experiments.  
For the sake of 
brevity, we report only cross-validation based power normalization, as 
square-rooting gives similar results.
In order to train LDA models, we first train a PLSA model,
and then fit Dirichlet priors on the topic-word and
document-topic distributions as inferred by PLSA.

\begin{figure}
\begin{center}
    \mysingleplot{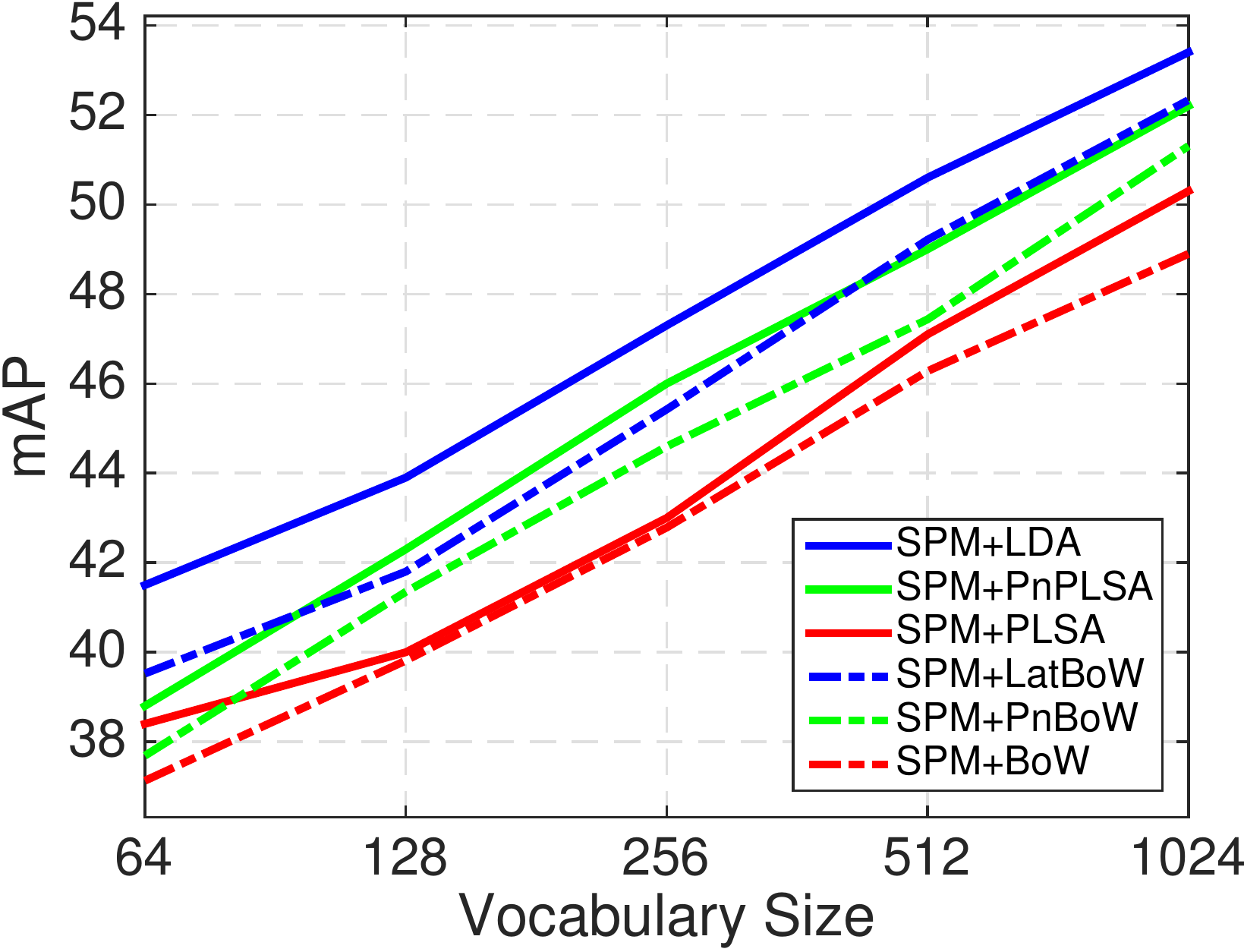}
\end{center}
    \caption{Topic models  ($T=2$, solid) compared with BoW models (dashed): BoW/PLSA (red), power-normalized BoW/PLSA (green), and P\'olya/LDA (blue). SPM grids are used in all experiments.}
\label{fig:VOC07_lda_overK_T2}
\end{figure}

\begin{figure}
\begin{center}
    \mysingleplot{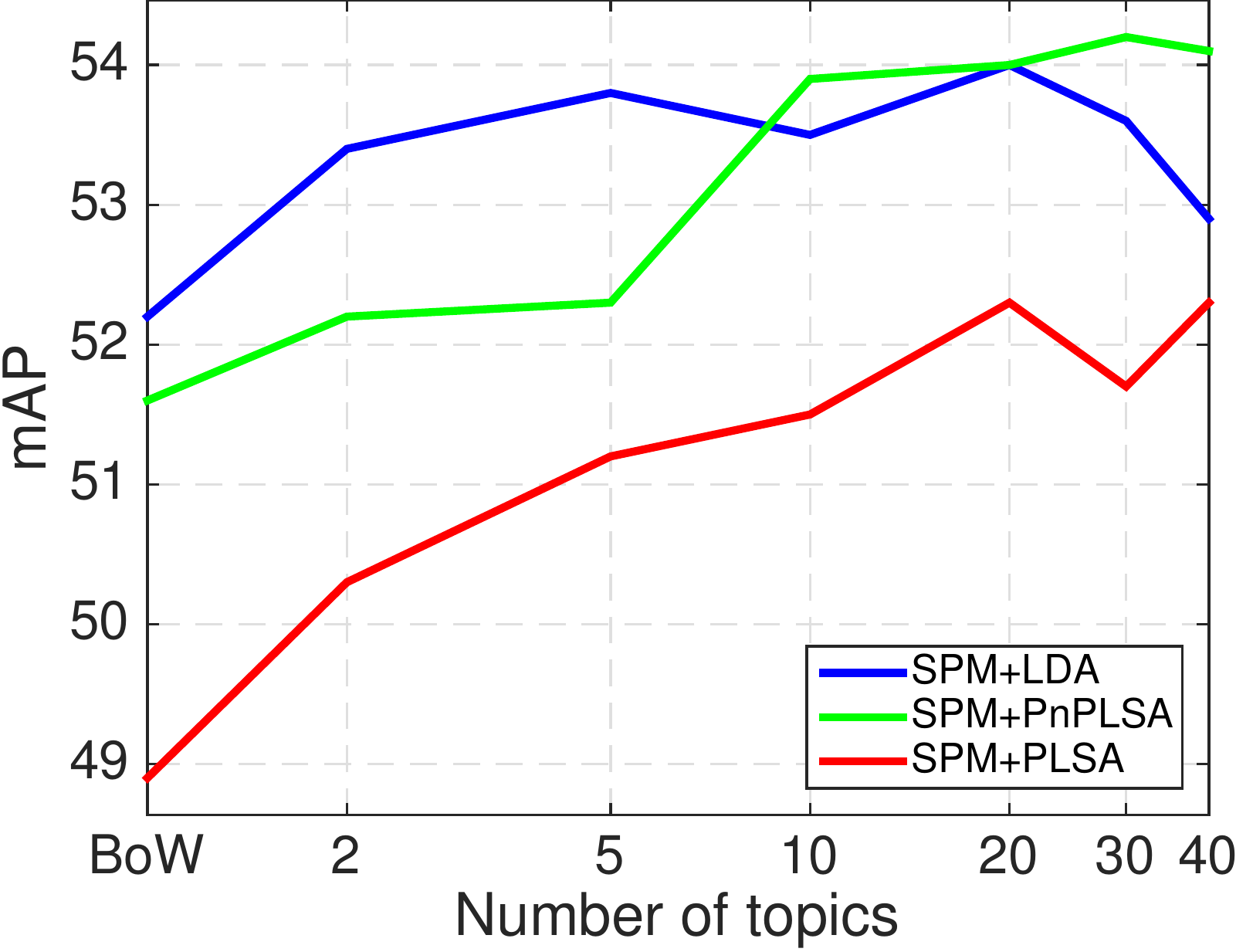}
\end{center}
    \caption{Performance when varying the number of topics:
PLSA (red), power-normalized PLSA (green), and LDA (blue). 
BoW/P\'olya model performance included as the left-most data point on each curve.
All experiments use SPM, and $K=1024$ visual words.
}
\label{fig:VOC07_lda_overT_K1024}
\end{figure}

In \fig{VOC07_lda_overK_T2},  we consider topic models  using $T\!=\!2$ topics for various dictionary sizes, and  
in \fig{VOC07_lda_overT_K1024} we use dictionaries of $K\!=\!1024$ visual
words, and consider performance as a function of the number of topics.

We observe that (i) topic models
consistently improve performance over BoW models, and (ii)  the plain
PLSA representations are consistently outperformed by the
power normalized version, and the LDA model. 
The LDA model requires less topics than (power-normalized) PLSA to obtain similar performance levels.
This is in line with our findings with the BoW model of the previous section.

%% file: experiments_mog.tex
\subsection{Evaluating latent MoG model}
\label{sec:xp:mog}

We now turn to the evaluation of the MoG-based image representations.
In order to speed-up the learning of the hyper-parameters, we fix the patch-to-word
soft-assignments as obtained from the MoG dictionary, and pre-compute the sufficient statistics of \Eq{latmog:sufstat} once.
We then iteratively update the model hyper-parameters, and the parameters of the posteriors on the per-image latent MoGs, as detailed in \sect{LatMoG}.

We initialize the Dirichlet distribution on the mixing weights by matching the moments of the distribution of normalized visual word frequencies $s_k^0$, which
gives an approximate maximum likelihood estimation~\cite{minka12report}.
Similarly, we initialize the hyper-parameters $a_k$ and $b_k$ of the Gamma prior on the precision of visual word $k$, by matching the mean and variance of empirical precision values computed from the sufficient statistics for each visual word, while weighting the contribution of each image by the count of visual word $k$ in that image. In this step, the empirical precision values of visual words with few associated descriptors can become too large and may lead to poor initialization. To deal with this issue, we truncate per-image empirical precision values with respect to the corresponding global empirical precision values scaled by a constant factor, which is cross-validated among a predefined set of values.
Finally, we initialize the hyper-parameters $m_k$ and $\beta_k$ by matching the mean and variance of the per-image empirical mean values computed from the sufficient statistics, again weighting each image by the count of visual word $k$ in that image.\footnote{Source code for LatMoG is available at \url{http://lear.inrialpes.fr/software}.}

In \tab{VOC07_gmmvslgmm_overK}, we compare representations based on Fisher vectors computed over MoG models, their two power normalized versions, and the latent MoG
model of \sect{LatMoG}. 
We can observe that the MoG representations
lead to better performance than the BoW and topic model
representations while using smaller vocabularies.
Furthermore, the discounting effect of power normalization and our latent variable model
 has a more pronounced effect here than
it has for  BoW models, improving mAP scores by around 4
points.  Also for the MoG models, our latent variable approach leads to  improvements
that are comparable to those obtained
by power normalization. So again, the benefits of
power normalization may be explained by using non-\iid latent
variable models that generate similar representations. 

\input{tables/voc07_mog}

\begin{figure}
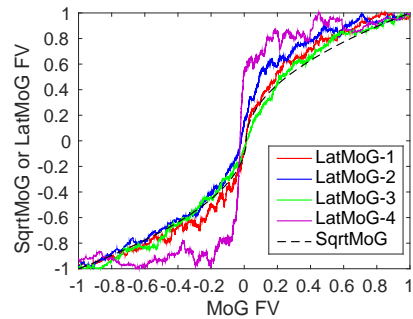

\begin{center}
\mysingleplot{learnt_curves/latmog_empirical_posneg}
\end{center}
\caption{Empirical comparison of components related to the Gaussian means of the power  normalized MoG FVs (SqrtMoG) and latent MoG FVs (LatMoG) \vs the non-power-normalized FV (horizontal axis). 
All FV values are scaled to the range [-1,1].}
\label{fig:LatMoGExampleCurves}
\end{figure}

Similar to \fig{PolyaExampleCurves}, we present an empirical comparison of the MoG FV and LatMoG FV based on a vocabulary of size $K=64$ components in \fig{LatMoGExampleCurves}. 
In this case we consider gradients \wrt the Gaussian mean parameters.
The transformation given by power normalization  is given for reference in dashed black.
Each LatMoG curve is obtained by sampling a dimension-cluster pair $(d,k)$, 
and plotting the LatMoG FV with respect to $m_{k,d}$ as a function of the MoG FV with respect to $\mu_{k,d}$ over different images. 
The  LatMoG curves are smoothed via a median filter for visualization purposes. 
We observe that the LatMoG model naturally generates FVs with discounting effects, as demonstrated by the curves similar to square-root transformation.
Note that the gradient in  \Eq{gradmean} for the LatMoG model is a joint function of the $s^0_k$, $s^1_k$ and $s^2_k$ statistics, which makes that plotting LatMoG FVs against MoG FVs results in non-smooth curves. 

%% file: tables/voc07_mog.tex
\begin{table}
\begin{center}
{\footnotesize
\input{autogen/mkAllFiguresAndTables/voc07_mog.tex}
}
\end{center}
\caption{Comparison of MoG-based FV representations: plain MoG,
      two types of power normalized MoG, and latent MoG.
}
\label{tab:VOC07_gmmvslgmm_overK}
\end{table}

%% file: autogen/mkAllFiguresAndTables/voc07_mog.tex
\begin{tabular}{cccccccc}
\hline
\multicolumn{1}{|c|}{SPM} & \multicolumn{1}{c||}{Method}  & \multicolumn{1}{c|}{32} & \multicolumn{1}{c|}{64} & \multicolumn{1}{c|}{128} & \multicolumn{1}{c|}{256} & \multicolumn{1}{c|}{512} & \multicolumn{1}{c|}{1024} \\
\hline
\noalign{\smallskip}
\hline
\multicolumn{1}{|c|}{No} & \multicolumn{1}{c||}{MoG} & \multicolumn{1}{c|}{49.1} & \multicolumn{1}{c|}{51.4} & \multicolumn{1}{c|}{53.1} & \multicolumn{1}{c|}{54.3} & \multicolumn{1}{c|}{55.0} & \multicolumn{1}{c|}{55.9}\\
\hline
\multicolumn{1}{|c|}{No} & \multicolumn{1}{c||}{SqrtMoG} & \multicolumn{1}{c|}{51.8} & \multicolumn{1}{c|}{54.7} & \multicolumn{1}{c|}{56.2} & \multicolumn{1}{c|}{58.2} & \multicolumn{1}{c|}{58.9} & \multicolumn{1}{c|}{60.2}\\
\hline
\multicolumn{1}{|c|}{No} & \multicolumn{1}{c||}{PnMoG} & \multicolumn{1}{c|}{\textbf{52.6}} & \multicolumn{1}{c|}{55.0} & \multicolumn{1}{c|}{\textbf{56.9}} & \multicolumn{1}{c|}{\textbf{59.0}} & \multicolumn{1}{c|}{\textbf{60.3}} & \multicolumn{1}{c|}{\textbf{61.1}}\\
\hline
\multicolumn{1}{|c|}{No} & \multicolumn{1}{c||}{LatMoG} & \multicolumn{1}{c|}{\textbf{52.9}} & \multicolumn{1}{c|}{\textbf{55.9}} & \multicolumn{1}{c|}{\textbf{56.6}} & \multicolumn{1}{c|}{\textbf{58.6}} & \multicolumn{1}{c|}{59.5} & \multicolumn{1}{c|}{60.2}\\
\hline
\noalign{\smallskip}
\hline
\multicolumn{1}{|c|}{Yes} & \multicolumn{1}{c||}{MoG} & \multicolumn{1}{c|}{53.1} & \multicolumn{1}{c|}{55.4} & \multicolumn{1}{c|}{56.2} & \multicolumn{1}{c|}{57.1} & \multicolumn{1}{c|}{57.4} & \multicolumn{1}{c|}{57.6}\\
\hline
\multicolumn{1}{|c|}{Yes} & \multicolumn{1}{c||}{SqrtMoG} & \multicolumn{1}{c|}{56.0} & \multicolumn{1}{c|}{57.9} & \multicolumn{1}{c|}{58.9} & \multicolumn{1}{c|}{60.3} & \multicolumn{1}{c|}{60.5} & \multicolumn{1}{c|}{60.8}\\
\hline
\multicolumn{1}{|c|}{Yes} & \multicolumn{1}{c||}{PnMoG} & \multicolumn{1}{c|}{56.6} & \multicolumn{1}{c|}{\textbf{58.4}} & \multicolumn{1}{c|}{\textbf{59.5}} & \multicolumn{1}{c|}{\textbf{61.1}} & \multicolumn{1}{c|}{\textbf{61.3}} & \multicolumn{1}{c|}{\textbf{61.8}}\\
\hline
\multicolumn{1}{|c|}{Yes} & \multicolumn{1}{c||}{LatMoG} & \multicolumn{1}{c|}{\textbf{57.3}} & \multicolumn{1}{c|}{\textbf{58.9}} & \multicolumn{1}{c|}{\textbf{59.4}} & \multicolumn{1}{c|}{60.4} & \multicolumn{1}{c|}{60.7} & \multicolumn{1}{c|}{60.7}\\
\hline
\end{tabular}

%% file: experiments_lhoodVsMAP.tex
\subsection{Relationship between model likelihood and
categorization performance}
\label{sec:xp:lhood}

We have seen that the Fisher vectors of our non-\iid image
models provide significantly better image classification performance compared to the Fisher
vectors of the corresponding \iid models, unless power normalization is used to implement a discounting transformation on the image
descriptors. In a broad sense, our experimental results
suggest that Fisher kernels combined with more powerful generative models can possibly lead to better image categorization performance. 

In order to investigate the relationship between the
image models and the categorization performance using the corresponding Fisher
vectors, we propose to empirically analyze the MoG models
and the corresponding image descriptors at a number of PCA
projection dimensions ($D$) and vocabulary sizes ($K$).
Here, we use the log-likelihood of each model on a
validation set as a measure of the generative power of
the models and evaluate the image categorization
performance of the corresponding Fisher vectors 
in terms of mAP scores on the \VOC 2007 dataset.

One important detail is that it may not be meaningful to
compare the image categorization performance across image
descriptors of different dimensionality: Our previous
experimental results have shown that the mAP scores
typically increase as the MoG Fisher vector descriptors
become higher dimensional.  
Therefore, we want to compare
the categorization performance across the image
descriptors of fixed dimensionality, \ie across the
$(D,K)$ pairs such that the product $D\times{K}$ is constant.  On
the other hand, the log-likelihood of MoG models are
comparable only if they operate in the same PCA
projection space. In order to overcome this difficulty,
we convert each pair of PCA and MoG models into a joint
generative model, which allows us to obtain comparable
log-likelihood values across different PCA subspaces.

We propose to obtain the joint generative models by
first defining a shared descriptor space as follows: Let 
$\phi(\x) = U^\text{T}(\x-\mu_0)$ be the full-dimensional PCA transformation function for the local descriptors, where
$\mathbf{\mu}_0$ is the empirical mean of the $D_0$-dimensional local
descriptors and $U$ is the $D_0{\times}D_0$ dimensional matrix of PCA basis column vectors. We note that $\phi(\x)$ does not apply dimension reduction, and the projection of a local descriptor $\x$ onto the $D$ dimensional PCA subspace is given by $\mathbf{I}_{D\times{D_0}} \phi(\x)$, \ie the first $D$ coordinates of $\phi(\x)$. Therefore, the density function of a given MoG model in the $D$-dimensional PCA subspace is given by
\begin{equation}
p(\x) = \sum_{k=1}^K \pi_k \mathcal{N}(\mathbf{I}_{D\times{D_0}} \phi(\x);\mathbf{\mu}_k,\Sigma_k).
\end{equation}
where $\mathbf{\pi}_k$ is the mixing weight,
$\mathbf{\mu}_k$ is the $D$-dimensional mean vector and
$\mathbf{\sigma}_k$ is the variances vector of the $k$-th component. Then, we can map the PCA dimension reduction
model and the MoG model into a new MoG model in the 
space of $\phi(\x)$ descriptors as follows:
\begin{equation} 
    p_0(\x) = \sum_k \pi_k \mathcal{N}( \phi(\x); \mathbf{\mu}_k^\prime, \mathbf{\sigma}_k^\prime)
\end{equation}
where each mean vector is defined as
\begin{equation}
    \mathbf{\mu}_k^\prime =  \mathbf{I}_{D_0\times{D}} \mathbf{\mu}_k ,
\end{equation}
and each variances vector $\mathbf{\sigma}_k^\prime$
is obtained by concatenating the corresponding
$D$-dimensional $\mathbf{\sigma}_k$ vector with
 the empirical global variances of the remaining $D_0-D$
 dimensions.

In our experiments, we have randomly sampled 300,000
SIFT descriptors to measure the average model log-likelihoods. We
evaluate the image categorization performance using
square-rooted and $\ell_2$ normalized MoG Fisher vectors,
without a spatial pyramid. We have utilized $(D,K)$ pairs
obtained by varying $D$ from 8 to 128 and $K$ from 64 to
4096.

\begin{figure}
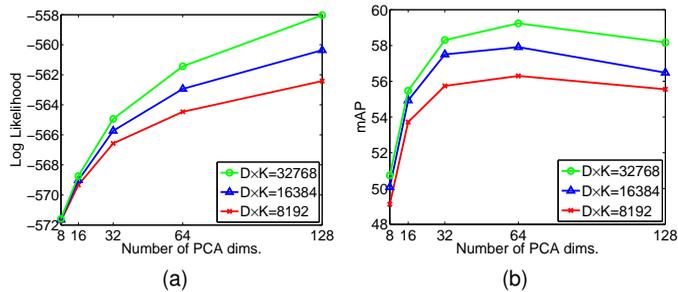

\begin{center}
    \subfloat[]{%
    \mydoubleplot{{journal/likelihood_vs_mAP/Jakob_2014_05_07/llk_0.01}.pdf}
    \label{fig:LoglkVsMAP:Loglk}
    }
    \subfloat[]{
    \mydoubleplot{{journal/likelihood_vs_mAP/Jakob_2014_05_07/mAP_0.01}.pdf}
    \label{fig:LoglkVsMAP:mAP}
    }
\end{center}
\caption{Evaluation of the model log-likelihood and the
    classification performance in terms of mAP scores as
    a function of the number of PCA dimensions ($D$) and
    the vocabulary size ($K$). The x-axis of each plot
    shows the number of PCA dimensions.  Each curve
    represents a set of $(D,K)$ values such that
$D\times{K}$ stays constant.}
\label{fig:LoglkVsMoG}
\end{figure}

\fig{LoglkVsMAP:Loglk} presents the model log-likelihood values and \fig{LoglkVsMAP:mAP}
presents the corresponding image classification mAP scores. The x-axis of each plot shows the number of PCA dimensions. Each curve represents a set of $(D,K)$ values where  $D\times{K}$ is constant. 
From the experimental results first we can see that increasing the number of PCA dimensions (and hence reducing the number of mixing components)  consistently increases the model log-likelihood. Second, the mAP scores similarly increase up to $D \leq 64$, but then degrade from $D=64$ to $D=128$. Therefore, even if the model log-likelihood and categorization performance are related,
they are not necessarily tightly correlated. Image categorization performance can be affected by several other factors, including the details of target categorization task, and  transformations applied to the Fisher vector representations, such as power and $\ell_2$ normalization here. 
Despite these findings, we believe that further investigation of the relationship between the modeling strength of generative models and the performance of the corresponding Fisher vectors for recognition tasks can lead to advances in unsupervised representation learning.

%% file: experiments_cnn.tex
\subsection{Experiments using CNN features}
\label{sec:xp:cnn}

We have so far utilized the SIFT local descriptors in our
experiments. In this section, we evaluate the latent MoG
representation based on local descriptors extracted using a
convolutional neural network (CNN) model~\cite{krizhevsky12nips}. 
For this purpose, we consider two feature
extraction schemes. First, we utilize the grid based
region sampling approach based on the work by Gong
\etal~\cite{gong14eccv} and Liu \etal~\cite{liu14nips},
and extract local descriptors by feeding cropped regions
to a CNN model.  Second, inspired from the R-CNN object
detector~\cite{girshick14cvpr}, we propose to extract local CNN
features for the image regions sampled by a candidate
window generation method. Unlike the R-CNN detector,
however, we utilize the region descriptors to extract
image descriptors using the Fisher kernel framework,
instead of evaluating individual regions as detection
candidates. To the best of our knowledge, we are first to
utilize detection proposals for this purpose.

In order to extract CNN features from regions sampled
on a grid, we follow the local region sampling approach
proposed by Liu \etal~\cite{liu14nips}: a given image is
first scaled to a size of $512\times512$ pixels, then,
regions of size $227\times227$ are sampled in a sliding
window fashion with a stride of $8$ pixels. This procedure results
in around $1300$ regions per image. The image patch
corresponding to each region sample is cropped and feed
into the CNN model of Krizhevsky \etal
\cite{krizhevsky12nips}, which is pre-trained on the
ImageNet ILSVRC2012 dataset~\cite{deng09cvpr} using the
Caffe library~\cite{jia14caffe}. Finally, the outputs of
the CNN model are used as the local descriptors.

In our second approach, we utilize the detection proposal regions
generated using the selective search method of
Uijlings \etal~\cite{uijlings13ijcv}. This method computes
multiple hierarchical segmentation trees for a given
image, and takes the segment bounding boxes as the detection
proposals.  This procedure results in around $1,500$
regions per image.  Following the R-CNN object detector,
we crop and re-size the window proposals to regions of
size $224\times224$, as required by the CNN model. 

As region descriptors we consider the layer six and seven activations of the
 CNN model. In order to speed up the
Fisher vector computations, we project the original
$4,096$-dimensional feature vectors to $128$ dimensions
using PCA. In our preliminary experiments, we have
verified that higher dimensional PCA projections does not
improve the image categorization performance.  Following
the \iid MoG based experiments in \cite{gong14eccv} and
\cite{liu14nips}, we use models with $K=100$ Gaussian
components, $\ell_2$ normalize the resulting image
representations, and do not use  SPM grids. 

\input{tables/voc07_cnn}

\input{tables/voc07_cnn_sota}

In \tab{VOC07_cnn}, we compare the MoG Fisher vector, its
power normalized versions, and the latent MoG Fisher vector
representations.  
 First, we observe that
using selective search regions for descriptor pooling
leads to consistently better results than using the grid
based regions. Given that both approaches use  a
comparable number of regions, the improvement using selective search regions is probably due to using regions of multiple scales, 
and  having a better object-to-clutter ratio.
Second, we observe that also in this setting using the Latent MoG model leads to improvements that are comparable to those
obtained by power normalization.
Third, best results are obtained with layer seven activations using power normalization and our latent model.

In \tab{VOC07_sota} we show that our results
are comparable to the recent results based
on a similar CNN models. The first row shows the
CNN baseline ($73.9\%$), as reported by Razavian
\etal~\cite{razavian14arxiv}, which corresponds to
training an SVM classifier over the full image CNN
descriptors. The same paper also shows that the
performance can be improved to $77.2\%$ by
applying feature transformations to image descriptors and
incorporating additional training examples via
transforming images. Bilen \etal~\cite{bilen14bmvc}
($80.9\%$) explicitly localizes object instances in images
using an iterative weakly supervised localization method.
The result shows that explicit localization of the objects
can help better categorization of the images. 
Liu \etal~\cite{liu14nips} ($76.9\%$) extract Fisher
vectors of a sparse coding based model over local CNN
features (see \app{varfv} for a detailed discussion of
their model).  Overall, we observe that our results using
power normalized MoG FVs ($78.0\%$) and latent MoG FVs
($77.1\%$) are comparable to the aforementioned recent
results, all of which are based on similar CNN models, and
validate the effectiveness of our Latent MoG model for
local feature aggregation.

We note that better results on the VOC'07 dataset have
recently been reported based on significantly different
CNN features and/or architectures. For example, Chatfield
\etal~\cite{chatfield14bmvc} achieve $82.4\%$ mAP by
utilizing the {\em OverFeat}~\cite{sermanet14iclr}
architecture, combined with a carefully selected set of
data augmentation, data normalization and CNN fine-tuning
techniques.  Wei \etal~\cite{wei14arxiv} achieve $85.2\%$
by max-pooling the class predictions over candidate
windows, utilizing additional training images, and using a
two-stage CNN fine-tuning approach. Simonyan and
Zisserman~\cite{simonyan14arxiv} report that the
classification performance can be improved up to $89.7\%$
mAP by using very deep network architectures, and
combining multiple CNN models. We can expect similar
improvements in the feature aggregation methods,
including ours, by utilizing these better-performing CNN
features.

\input{tables/MITIndoor_cnn}

In order to validate our results on a second dataset, we
evaluate our latent MoG model on the MIT Indoor dataset.
The dataset contains 6,700 images, each of which is
labeled with one of the 67 indoor scene categories.  
Before extracting window proposals, 
we resize each image such that the larger dimension is 500 pixels.
We use the standard split for the dataset, which provides 80 train and 20 test images per
class, and evaluate the results in terms of average
classification accuracy.

The results for  MIT Indoor are presented
\tab{MITIndoor_cnn}. In each row, we evaluate a combination
of the 6-th or 7-th CNN layer with the grid based or
selective search based regions. Our results are overall
consistent with those we obtain on VOC 2007: (i) using
selective search regions leads to better performance, and (ii) using the Latent MoG model
leads to significant improvements, comparable to those obtained by power normalization. Therefore,
the results again support that  the benefits of
power normalization can be explained by their similarity to non-\iid latent
variable models that generate similar transformations. 

Finally, in \tab{MITIndoor_sota}, we compare our results
on the MIT Indoor dataset with the state-of-the-art. The
first methods, Juneja \etal~\cite{juneja13cvpr} ($63.2\%$)
and Doersch \etal~\cite{doersch13nips} ($64.0\%$), extract
mid-level representations by explicitly localizing
discriminative image regions. In the next two rows, we
observe that the CNN baseline improves from $58.4\%$
to $69.0\%$ using the feature and image transformations
proposed by Razavian \etal~\cite{razavian14arxiv}.  The
sparse coding Fisher vectors proposed by Liu
\etal~\cite{liu14nips} result in a comparable performance
at $68.2\%$. Gong \etal~\cite{gong14eccv} ($68.9\%$)
utilizes power normalized VLAD features over the CNN descriptors
extracted from multi-scale grid-based regions, in
combinations with the full image CNN features.  Overall,
we observe that our approach
using power normalized MoG FVs ($69.1\%$) and latent MoG FVs
($69.1\%$) over selective search regions provide
state-of-the-art performance on the MIT Indoor dataset.

\input{tables/MITIndoor_cnn_sota}

%% file: tables/voc07_cnn.tex
\begin{table}
\begin{center}
{\footnotesize
\input{autogen/mkAllFiguresAndTables/voc07_cnn.tex}
}
\end{center}
\caption{Comparison of mAP scores on PASCAL VOC'07 dataset: plain MoG,
two types of power normalized MoG and latent MoG.} 
\label{tab:VOC07_cnn}
\end{table}

%% file: autogen/mkAllFiguresAndTables/voc07_cnn.tex
\begin{tabular}{cccccc}
\hline
\multicolumn{1}{|c|}{Regions} & \multicolumn{1}{c||}{CNN Layer}  & \multicolumn{1}{c|}{MoG} & \multicolumn{1}{c|}{SqrtMoG} & \multicolumn{1}{c|}{PnMoG} & \multicolumn{1}{c|}{LatMoG} \\
\hline
\noalign{\smallskip}
\hline
\multicolumn{1}{|c|}{Grid}      & \multicolumn{1}{c||}{fc6}  & \multicolumn{1}{c|}{69.4} & \multicolumn{1}{c|}{\textbf{74.1}} & \multicolumn{1}{c|}{\textbf{74.3}} & \multicolumn{1}{c|}{73.3}\\
\hline
\multicolumn{1}{|c|}{Grid}      & \multicolumn{1}{c||}{fc7}  & \multicolumn{1}{c|}{66.6} & \multicolumn{1}{c|}{74.6} & \multicolumn{1}{c|}{\textbf{75.7}} & \multicolumn{1}{c|}{\textbf{75.3}}\\
\hline
\multicolumn{1}{|c|}{Selective} & \multicolumn{1}{c||}{fc6}  & \multicolumn{1}{c|}{74.2} & \multicolumn{1}{c|}{76.8} & \multicolumn{1}{c|}{\textbf{77.0}} & \multicolumn{1}{c|}{75.5}\\
\hline
\multicolumn{1}{|c|}{Selective} & \multicolumn{1}{c||}{fc7}  & \multicolumn{1}{c|}{74.5} & \multicolumn{1}{c|}{77.8} & \multicolumn{1}{c|}{\textbf{78.0}} & \multicolumn{1}{c|}{77.1}\\
\hline
\end{tabular}

%% file: tables/voc07_cnn_sota.tex
\begin{table}
\begin{center}
{\footnotesize
\begin{tabular}{|c|c|}
\hline
Method & mAP \\
\hline
\noalign{\smallskip}
\hline
CNN baseline~\cite{razavian14arxiv}    & 73.9 \\
\hline
Razavian \etal~\cite{razavian14arxiv}        & 77.2 \\
\hline
Bilen \etal~\cite{bilen14bmvc}               & {\bf 80.9} \\
\hline
Liu \etal~\cite{liu14nips}                   & 76.9 \\
\hline
\noalign{\smallskip}
\hline
Ours (PnMoG, sel.\ search, fc7)                              & 78.0 \\
\hline
Ours (LatMoG, sel.\ search, fc7)                               & 77.1 \\
\hline
\end{tabular}
}
\end{center}
\caption{Comparison of the power normalized MoG and latent MoG
    representations against recent results on the PASCAL
    VOC'07 dataset. } 
\label{tab:VOC07_sota}
\end{table}

%% file: tables/MITIndoor_cnn.tex
\begin{table}
\begin{center}
{\footnotesize
\input{autogen/mkAllFiguresAndTables/MITIndoor_cnn.tex}
}
\end{center}
\caption{Comparison of classification accuracy on MIT Indoor: plain MoG,
two types of power normalized MoG and latent MoG.} 
\label{tab:MITIndoor_cnn}
\end{table}

%% file: autogen/mkAllFiguresAndTables/MITIndoor_cnn.tex
\begin{tabular}{cccccc}
\hline
\multicolumn{1}{|c|}{Regions} & \multicolumn{1}{c||}{CNN Layer}  & \multicolumn{1}{c|}{MoG} & \multicolumn{1}{c|}{SqrtMoG} & \multicolumn{1}{c|}{PnMoG} & \multicolumn{1}{c|}{LatMoG} \\
\hline
\noalign{\smallskip}
\hline
\multicolumn{1}{|c|}{Grid}      & \multicolumn{1}{c||}{fc6}  & \multicolumn{1}{c|}{60.1} & \multicolumn{1}{c|}{66.0} & \multicolumn{1}{c|}{\textbf{67.3}} & \multicolumn{1}{c|}{62.2}\\
\hline
\multicolumn{1}{|c|}{Grid}      & \multicolumn{1}{c||}{fc7}  & \multicolumn{1}{c|}{57.0} & \multicolumn{1}{c|}{\textbf{64.8}} & \multicolumn{1}{c|}{\textbf{65.0}} & \multicolumn{1}{c|}{61.5}\\
\hline
\multicolumn{1}{|c|}{Selective} & \multicolumn{1}{c||}{fc6}  & \multicolumn{1}{c|}{66.6} & \multicolumn{1}{c|}{\textbf{69.4}} & \multicolumn{1}{c|}{\textbf{69.7}} & \multicolumn{1}{c|}{68.2}\\
\hline
\multicolumn{1}{|c|}{Selective} & \multicolumn{1}{c||}{fc7}  & \multicolumn{1}{c|}{65.2} & \multicolumn{1}{c|}{\textbf{69.0}} & \multicolumn{1}{c|}{\textbf{69.1}} & \multicolumn{1}{c|}{\textbf{69.1}}\\
\hline
\end{tabular}

%% file: tables/MITIndoor_cnn_sota.tex
\begin{table}
\begin{center}
{\footnotesize
\begin{tabular}{|c|c|}
\hline
Method & Accuracy \\
\hline
\noalign{\smallskip}
\hline
Juneja \etal~\cite{juneja13cvpr}               & 63.2 \\
\hline
Doersch \etal~\cite{doersch13nips}             & 64.0 \\
\hline
CNN baseline~\cite{razavian14arxiv}            & 58.4  \\
\hline
Razavian \etal~\cite{razavian14arxiv}          & 69.0  \\
\hline
Liu \etal~\cite{liu14nips}                     & 68.2  \\
\hline
Gong \etal~\cite{gong14eccv}                   & 68.9  \\
\hline
\noalign{\smallskip}
\hline
Ours (PnMoG, sel.\ search, fc7)              & {\bf 69.1}  \\
\hline
Ours (LatMoG, sel.\ search, fc7)              & {\bf 69.1} \\
\hline
\end{tabular}
}
\end{center}
\caption{Comparison of the power normalized MoG and latent MoG
    representations against recent results on the MIT
    Indoor dataset. } 
\label{tab:MITIndoor_sota}
\end{table}

%% file: conclusion.tex
\section{Conclusions}
\label{sec:conclusion}

In this paper we have introduced latent variable models
for local image descriptors, which avoid the common but
unrealistic \iid assumption. The Fisher vectors of our
non-\iid models are functions computed from the same
sufficient statistics as those used to compute Fisher
vectors of the corresponding \iid models.  In fact, these
functions are similar to transformations that have been
used in earlier work in an ad-hoc manner, such as the
power normalization, or signed-square-root.  Our models provide an explanation of the
success of such transformations, since  we derive them
here by removing the unrealistic \iid assumption from the
popular BoW and MoG models.  Second, we have shown that
gradients of the variational free-energy bound on the
log-likelihood gives exact Fisher score vectors as long as
the variational posterior distribution is exact. Third, we
have shown that approximate Fisher vectors for the
proposed latent MoG model can be successfully extracted
using the variational Fisher vector framework. Finally, we
have shown that the Fisher vectors of our non-\iid MoG
model over CNN region descriptors extracted on selectively
sampled windows lead to image categorization performance
that is comparable or superior to that obtained with
state-of-the-art feature aggregation representations based on \iid
models.

%% file: appendices.tex
\appendix[A. Variational Fisher kernel examples]
\label{app:varfv}

In this section, we give two examples that
illustrate  applications of the variational
FV framework, in addition to the models considered in the main text. 


In our first example, we derive a fast variant of the MoG FV representation using the 
variational Fisher kernel formulation.  
Recall that the final MoG FV image
representation is obtained by aggregating
$K(1+2D)$-dimensional per-patch FVs. Therefore,
the cost of feature extraction grows linearly with respect
to $K$, $D$ and $N$. One way to speed up this process,
without sacrificing the descriptor 
dimensionality, is to  hard-assign each local
descriptor to visual word with the highest posterior
probability.  Using hard-assignment, each local descriptor
produces a $(1+2D)$ dimensional descriptor, therefore, the
aggregation speeds-up by a factor of $K$.  As noted in
\cite{sanchez13ijcv}, the MoG FV descriptor in
this case can be also interpreted as a generalization of
the VLAD descriptor \cite{jegou11pami}.

Although the hard-assignment method can provide significantly speeds
up in the descriptor aggregation process, it may also
cause significant information loss \cite{gemert10pami}.
This problem can be addressed by utilizing {\em clipped}
posterior weights within the variational FV
framework. More specifically, we can define the family of approximate posteriors
$\mathcal Q$ as those distributions with  at most $K^\prime$ non-zero values. 
The best approximation to a given $p(k|x)$ is then obtained by re-normalizing the largest  $K^\prime$ values of $p(k|x)$ and setting the other values to zero.
In this case, each patch yields a descriptor with at most
$K^\prime(1+2D)$ non-zero values, which translates into a
aggregation speed up of factor $\frac{K}{K^\prime}$.
The number of non-zeros $K^\prime$ can be set to strike a balance between the information loss and the aggregation cost. 
This shows that clipping the posteriors to speed-up the computation of FVs, as \eg done in \cite{cinbis13iccv}, can be justified in the variational framework. 
The MoG model can also be learned in a coherent manner, by optimizing the obtained variational bound instead of the log-likelihood. This forces the MoG components to be more separated, so that the true posteriors will concentrate on few components.


As a second example, we show that the derivation of the
sparse coding FVs of Liu \etal~\cite{liu14nips}, which we
have experimentally compared against in \sect{xp:cnn}, can
be significantly simplified using the variational
formulation. In their approach, a $D$-dimensional local
descriptor $\x$ is modeled by a mixture of basis vectors: 
\bea
p(\x) = \int p(\x|\u;\B) p(\u) d\u
\label{eq:liu:mixture}
\eea
where $\u$ is the latent vector of mixing weights of length
$K$, and $\B$ is the dictionary matrix with each of the $K$ columns corresponding to a $D$-dimensional basis vector.
The distribution $p(\x|\u;\B)$ is a Gaussian  with mean $\B\u$,
and  covariance matrix equal to a multiple of the identity matrix, and $p(\u)$ is
the Laplacian prior on the mixture weights. Liu
\etal~\cite{liu14nips} propose to approximate $p(\x)$ by
the point estimate for $\u$ that maximizes the likelihood:
\bea
p(\x) \approx p(\x|\u^\star;\B) p(\u^\star)
%
\label{eq:liu:pointestimate:p}
\eea
where \bea
\u^\star = \argmax_\u p(\x|\u;\B)p(\u) .
\label{eq:liu:pointestimate:u}
\eea
In order to compute FVs for this model, we need
to compute the gradients of \Eq{liu:pointestimate:p} with
respect to the dictionary matrix $\B$. However, as noted
in \cite{liu14nips}, this is leads to a relatively
complicated calculation since $\u^\star$ is dependent on $\B$. Using a series of techniques, it is shown in
\cite{liu14nips} that the gradient is given by:
\bea
\grad{\log p(\x)}{\B} = (\x-\B\u^\star)\u^\star
\label{eq:liu:grad}
\eea
Instead, we can use the variational Fisher kernel formulation
to achieve the same result in a simpler way. 
For this purpose, we define the class of approximate posteriors $\mathcal Q$ as the set of delta peaks that put all mass at a single value $\u$. 
It is then easy to see that the optimal $q\in \mathcal Q$ that maximizes the variational bound is  $q(\u^\star)=1$ and $q(\u\neq\u^\star)=0$. Given the optimal $q$, the variational FV is given by:
\bea
\grad{F}{\B} & = &  \grad{\ex{q}{\ln p(\x,\u)}}{\B} \\
             & = &  \grad{\ln p(\x,\u^\star)}{\B}
\eea
Compared to \Eq{liu:pointestimate:u}, this is a much
simpler derivative operation since the gradient is now
decoupled from the $\u^\star$ estimation step. It can be
easily shown that the resulting gradient is equivalent to
\Eq{liu:grad}.  This shows that the variational
FV formulation can  be preferable
over the original FV formulation.

\smallskip

\noindent {\bf Acknowledgements.} This work was supported by the
European integrated project AXES and the ERC advanced grant ALLEGRO.

%% file: bios.tex
\begin{IEEEbiography}[{\includegraphics[width=1in,height=1.25in,clip,keepaspectratio]{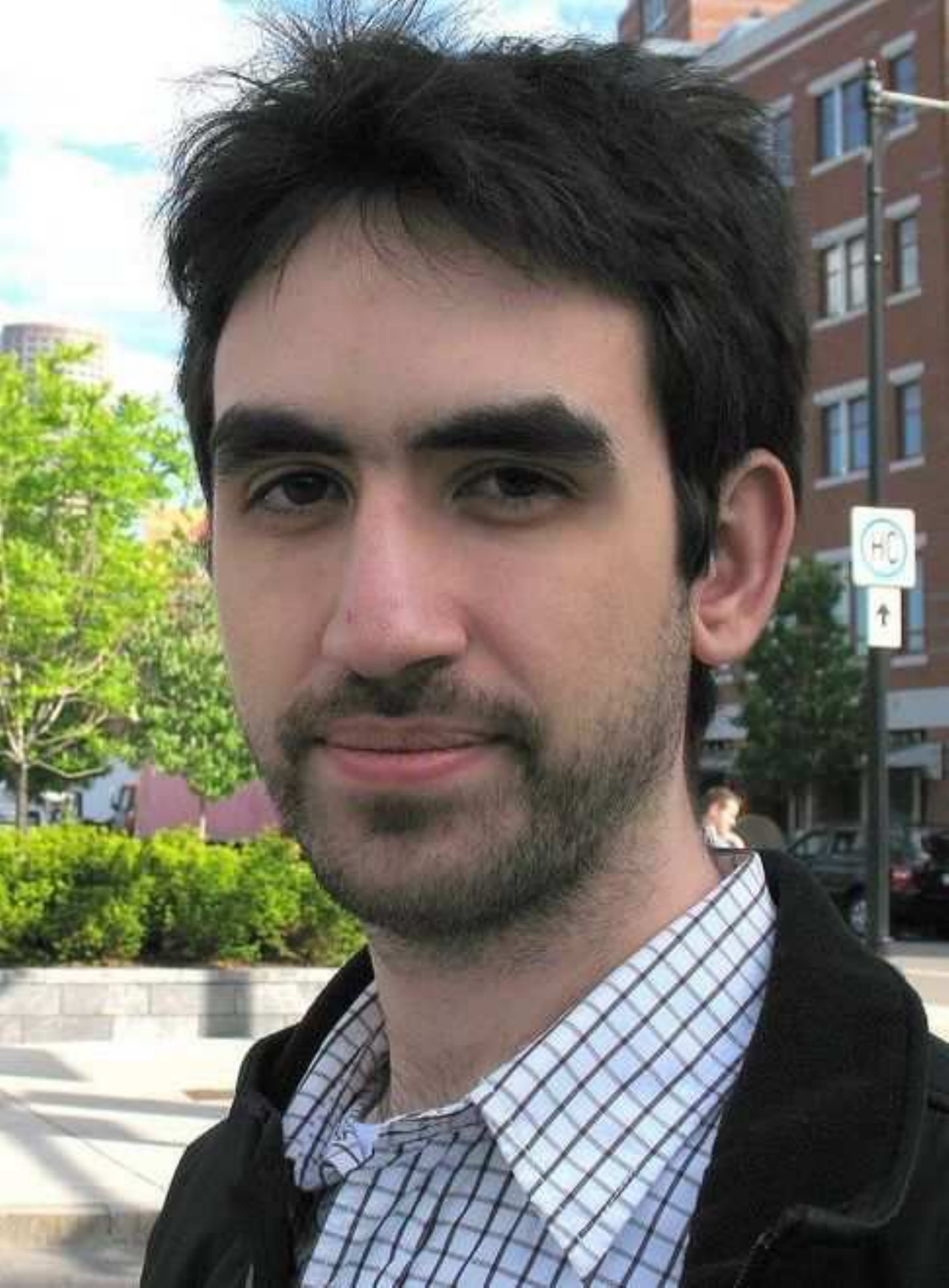}}]{Ramazan Gokberk Cinbis}
graduated from Bilkent University, Turkey, in 2008, and
received an M.A. degree in computer science from Boston University, USA, in 2010.
He was
a doctoral student in the LEAR team, at INRIA Grenoble,
France, from 2010 until 2014, and received a PhD degree in
computer science from Universit\'{e} de Grenoble, France,
in 2014. 
His research interests include computer vision and machine learning.
\end{IEEEbiography}

\begin{IEEEbiography}[{\includegraphics[width=1in,height=1.25in,clip,keepaspectratio]{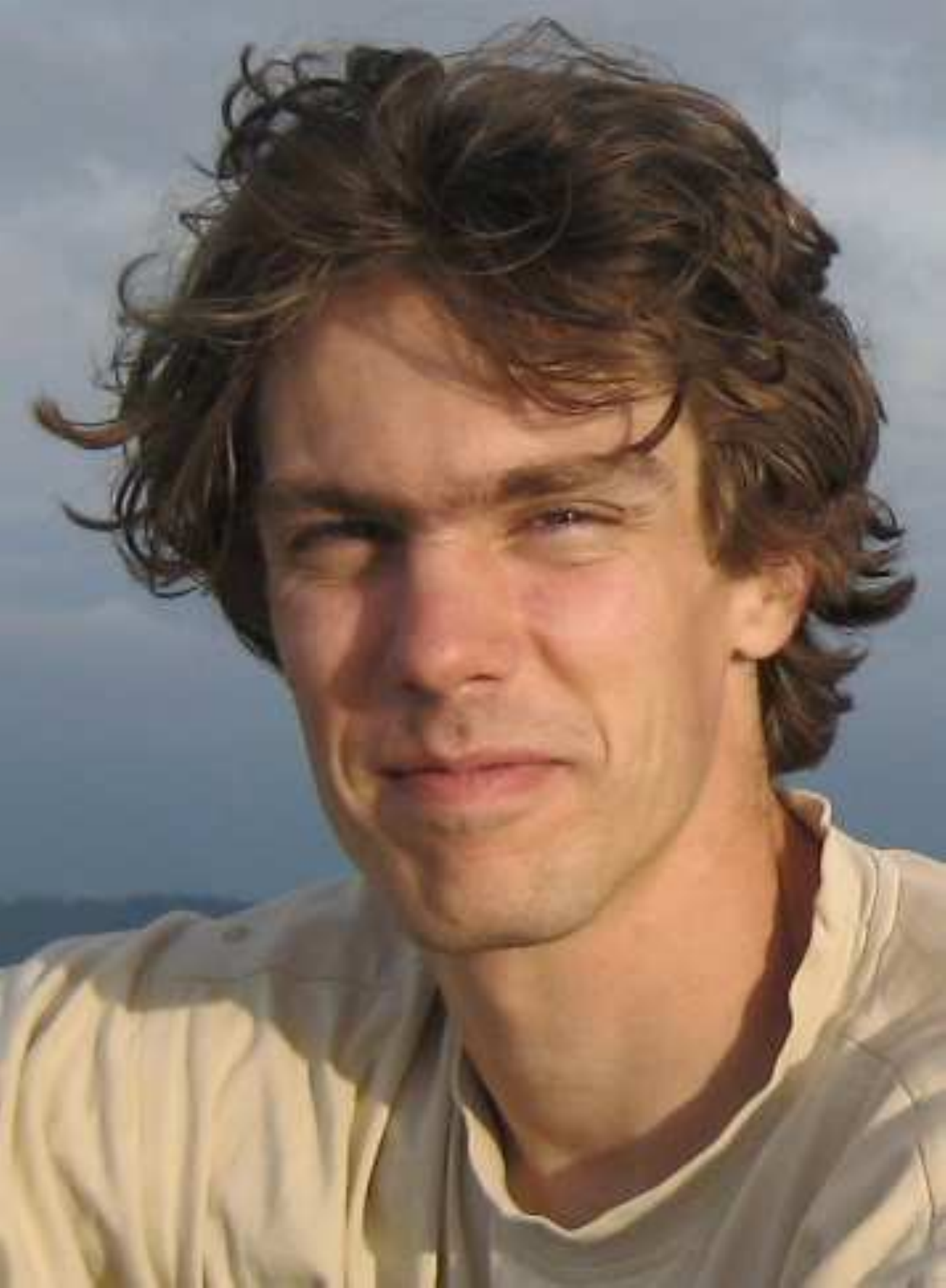}}]{Jakob Verbeek}
received a PhD degree in computer science in 2004 from the University of Amsterdam, The Netherlands. 
After being a postdoctoral researcher at the University of Amsterdam and at INRIA Rh\^one-Alpes, he has
been a full-time researcher  at INRIA, Grenoble, France, since 2007. 
His research interests include machine learning and computer vision, with special interest in applications of
statistical models in computer vision. 
\end{IEEEbiography}

\begin{IEEEbiography}[{\includegraphics[width=1in,height=1.25in,clip,keepaspectratio]{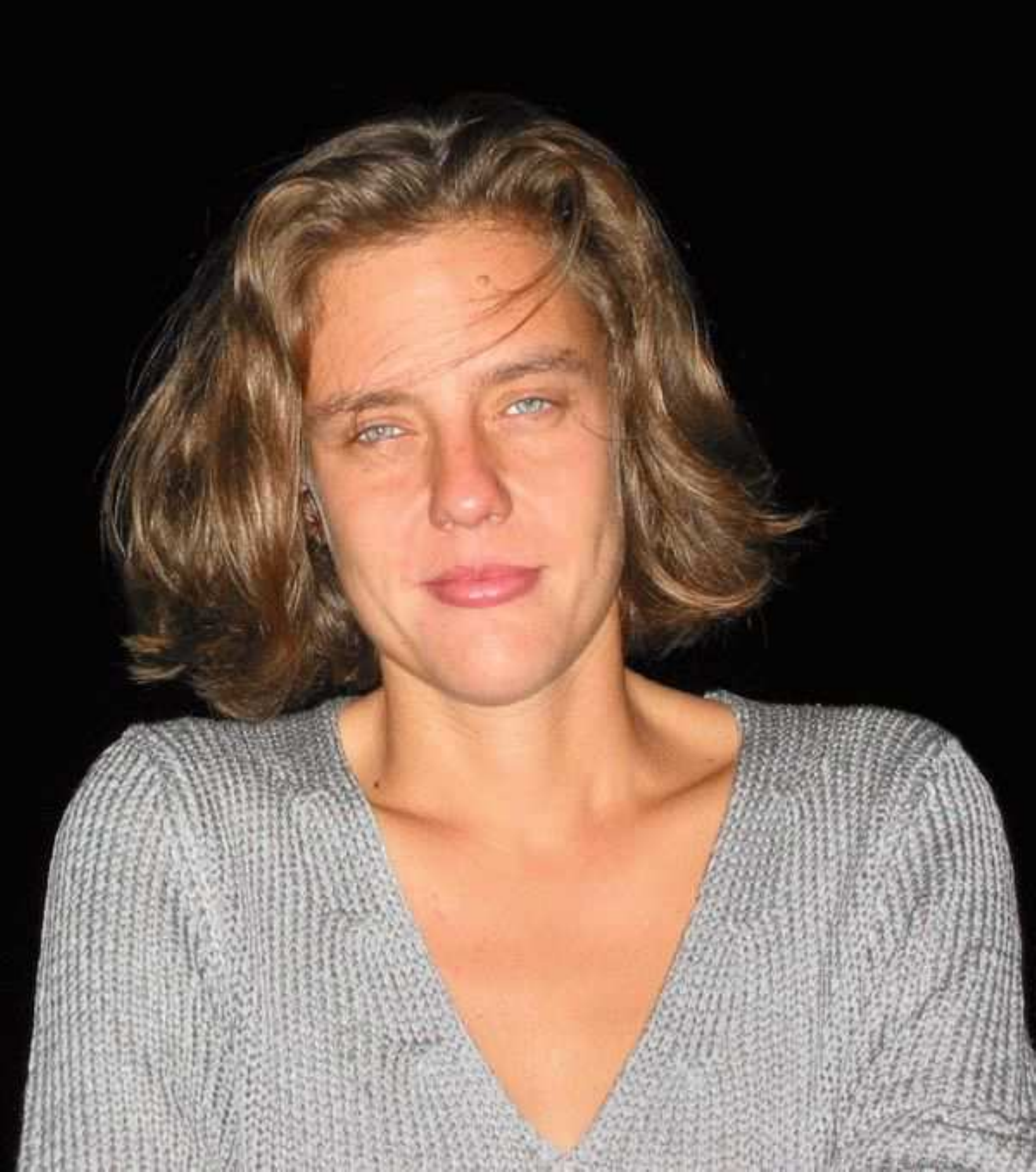}}]{Cordelia Schmid}
holds a M.S. degree in computer science
from the University of Karlsruhe and a doctorate from the
Institut National Polytechnique de Grenoble. She is a
research director at INRIA Grenoble where she directs the
LEAR team. In 2006 and 2014, she was awarded the Longuet-Higgins
prize for fundamental contributions in computer vision
that have withstood the test of time. In 2012, she
obtained an ERC advanced grant. She is a fellow of IEEE.
\end{IEEEbiography}

%% file: paper.bbl
\begin{thebibliography}{10}\itemsep=-1pt

\bibitem{bilen14bmvc}
H.~Bilen, M.~Pedersoli, and T.~Tuytelaars.
\newblock Weakly supervised object detection with posterior regularization.
\newblock In {\em British Machine Vision Conference}, 2014.

\bibitem{bishop06patrec}
C.~Bishop.
\newblock {\em Pattern recognition and machine learning}.
\newblock {Spinger-Verlag}, 2006.

\bibitem{blei03jmlr}
D.~Blei, A.~Ng, and M.~Jordan.
\newblock Latent {D}irichlet allocation.
\newblock {\em Journal of Machine Learning Research}, 3:993--1022, 2003.

\bibitem{chandalia06nipsw}
G.~Chandalia and M.~Beal.
\newblock Using {Fisher} kernels from topic models for dimensionality
  reduction.
\newblock In {\em NIPS Workshop on Novel Applications of Dimensionality
  Reduction}, 2006.

\bibitem{chappelier09mlkdd}
J.~C. Chappelier and E.~Eckard.
\newblock {PLSI}: The true {Fisher} kernel and beyond.
\newblock In {\em Machine Learning and Knowledge Discovery in Databases}, pages
  195--210. Springer, 2009.

\bibitem{chatfield11bmvc}
K.~Chatfield, V.~Lempitsky, A.~Vedaldi, and A.~Zisserman.
\newblock The devil is in the details: an evaluation of recent feature encoding
  methods.
\newblock In {\em British Machine Vision Conference}, 2011.

\bibitem{chatfield14bmvc}
K.~Chatfield, K.~Simonyan, A.~Vedaldi, and A.~Zisserman.
\newblock Return of the devil in the details: Delving deep into convolutional
  nets.
\newblock In {\em British Machine Vision Conference}, 2014.

\bibitem{cinbis12cvpr}
R.~G. Cinbis, J.~Verbeek, and C.~Schmid.
\newblock Image categorization using {F}isher kernels of non-iid image models.
\newblock In {\em IEEE Conference on Computer Vision and Pattern Recognition},
  2012.

\bibitem{cinbis13iccv}
R.~G. Cinbis, J.~Verbeek, and C.~Schmid.
\newblock Segmentation driven object detection with {F}isher vectors.
\newblock In {\em International Conference on Computer Vision}, 2013.

\bibitem{dance04eccv}
G.~Csurka, C.~Dance, L.~Fan, J.~Willamowski, and C.~Bray.
\newblock Visual categorization with bags of keypoints.
\newblock In {\em ECCV Int. Workshop on Stat. Learning in Computer Vision},
  2004.

\bibitem{deng09cvpr}
J.~Deng, W.~Dong, R.~Socher, L.-J. Li, K.~Li, and L.~Fei-Fei.
\newblock Imagenet: A large-scale hierarchical image database.
\newblock In {\em IEEE Conference on Computer Vision and Pattern Recognition},
  2009.

\bibitem{doersch13nips}
C.~Doersch, A.~Gupta, and A.~A. Efros.
\newblock Mid-level {Visual} {Element} {Discovery} as {Discriminative} {Mode}
  {Seeking}.
\newblock In {\em Advances in Neural Information Processing Systems}, pages
  494--502, 2013.

\bibitem{elkan05spire}
C.~Elkan.
\newblock Deriving {TF-IDF} as a {F}isher kernel.
\newblock In {\em String Processing and Information Retrieval}, 2005.

\bibitem{everingham15ijcv}
M.~Everingham, S.~M.~A. Eslami, L.~Van~Gool, C.~K.~I. Williams, J.~Winn, and
  A.~Zisserman.
\newblock The {Pascal} {Visual} {Object} {Classes} {Challenge}.
\newblock {\em International Journal on Computer Vision}, 111(1):98--136, Jan.
  2015.

\bibitem{girshick14cvpr}
R.~Girshick, J.~Donahue, T.~Darrell, and J.~Malik.
\newblock Rich feature hierarchies for accurate object detection and semantic
  segmentation.
\newblock In {\em IEEE Conference on Computer Vision and Pattern Recognition},
  2013.

\bibitem{gong14eccv}
Y.~Gong, L.~Wang, R.~Guo, and S.~Lazebnik.
\newblock Multi-scale {Orderless} {Pooling} of {Deep} {Convolutional}
  {Activation} {Features}.
\newblock In {\em European Conference on Computer Vision}, 2014.

\bibitem{halevy09is}
A.~Halevy, P.~Norvig, and F.~Pereira.
\newblock The unreasonable effectiveness of data.
\newblock {\em Intelligent Systems, {IEEE}}, 24(2):8--12, 2009.

\bibitem{hofmann99nips}
T.~Hofmann.
\newblock Learning the similarity of documents: An information-geometric
  approach to document retrieval and categorization.
\newblock In {\em Advances in Neural Information Processing Systems}, pages
  914--920, 1999.

\bibitem{hofmann01ml}
T.~Hofmann.
\newblock Unsupervised learning by probabilistic latent semantic analysis.
\newblock {\em Machine Learning}, 42(1/2):177--196, 2001.

\bibitem{jaakkola99nips}
T.~Jaakkola and D.~Haussler.
\newblock Exploiting generative models in discriminative classifiers.
\newblock In {\em Advances in Neural Information Processing Systems}, 1999.

\bibitem{jegou09cvpr}
H.~J\'egou, M.~Douze, and C.~Schmid.
\newblock On the burstiness of visual elements.
\newblock In {\em IEEE Conference on Computer Vision and Pattern Recognition},
  2009.

\bibitem{jegou11pami}
H.~J{\'e}gou, F.~Perronnin, M.~Douze, J.~S{\'a}nchez, P.~P{\'e}rez, and
  C.~Schmid.
\newblock Aggregating local image descriptors into compact codes.
\newblock {\em IEEE Transactions on Pattern Analysis and Machine Intelligence},
  34(9):1704--1716, 2012.

\bibitem{jia14caffe}
Y.~Jia, E.~Shelhamer, J.~Donahue, S.~Karayev, J.~Long, R.~Girshick,
  S.~Guadarrama, and T.~Darrell.
\newblock Caffe: Convolutional architecture for fast feature embedding.
\newblock {\em arXiv preprint arXiv:1408.5093}, 2014.

\bibitem{jordan99ml}
M.~Jordan, Z.~Ghahramani, T.~Jaakola, and L.~Saul.
\newblock An introduction to variational methods for graphical models.
\newblock {\em Machine Learning}, 37(2):183--233, 1999.

\bibitem{juneja13cvpr}
M.~Juneja, A.~Vedaldi, C.~V. Jawahar, and A.~Zisserman.
\newblock Blocks that shout: Distinctive parts for scene classification.
\newblock In {\em IEEE Conference on Computer Vision and Pattern Recognition},
  2013.

\bibitem{kobayashi14cvpr}
T.~Kobayashi.
\newblock Dirichlet-based histogram feature transform for image classification.
\newblock In {\em IEEE Conference on Computer Vision and Pattern Recognition},
  2014.

\bibitem{krapac11iccv}
J.~Krapac, J.~Verbeek, and F.~Jurie.
\newblock Modeling spatial layout with {F}isher vectors for image
  categorization.
\newblock In {\em International Conference on Computer Vision}, 2011.

\bibitem{krizhevsky12nips}
A.~Krizhevsky, I.~Sutskever, and G.~Hinton.
\newblock Imagenet classification with deep convolutional neural networks.
\newblock In {\em Advances in Neural Information Processing Systems}, 2012.

\bibitem{larlus09ivc}
D.~Larlus and F.~Jurie.
\newblock Latent mixture vocabularies for object categorization and
  segmentation.
\newblock {\em {I}mage and {V}ision {C}omputing}, 27(5):523--534, 2009.

\bibitem{lazebnik06cvpr}
S.~Lazebnik, C.~Schmid, and J.~Ponce.
\newblock Beyond bags of features: spatial pyramid matching for recognizing
  natural scene categories.
\newblock In {\em IEEE Conference on Computer Vision and Pattern Recognition},
  2006.

\bibitem{liu14nips}
L.~Liu, C.~Shen, L.~Wang, A.~van~den Hengel, and C.~Wang.
\newblock Encoding high dimensional local features by sparse coding based
  {Fisher} vectors.
\newblock In {\em Advances in Neural Information Processing Systems}, 2014.

\bibitem{mackay03book}
D.~J. MacKay.
\newblock {\em Information theory, inference, and learning algorithms}.
\newblock Cambridge University Press, 2003.

\bibitem{madsen05icml}
R.~Madsen, D.~Kauchak, and C.~Elkan.
\newblock Modeling word burstiness using the {D}irichlet distribution.
\newblock In {\em International Conference on Machine Learning}, 2005.

\bibitem{minka12report}
T.~Minka.
\newblock Estimating a {Dirichlet} distribution.
\newblock
  \nolinkurl{http://research.microsoft.com/en-us/um/people/minka/papers/dirichlet/minka-dirichlet.pdf},
  2012.

\bibitem{perina09nips}
A.~Perina, M.~Cristani, U.~Castellani, V.~Murino, and N.~Jojic.
\newblock Free energy score space.
\newblock In {\em Advances in Neural Information Processing Systems}, 2009.

\bibitem{perina15pami}
A.~Perina and N.~Jojic.
\newblock Capturing spatial interdependence in image features: the counting
  grid, an epitomic representation for bags of features.
\newblock {\em IEEE Transactions on Pattern Analysis and Machine Intelligence},
  2015.

\bibitem{perina14icpr}
A.~Perina, M.~Kesa, and M.~Bicego.
\newblock Expression microarray data classification using counting grids and
  {Fisher} kernel.
\newblock In {\em IAPR International Conference on Pattern Recognition}, 2014.

\bibitem{perronnin07cvpr}
F.~Perronnin and C.~Dance.
\newblock Fisher kernels on visual vocabularies for image categorization.
\newblock In {\em IEEE Conference on Computer Vision and Pattern Recognition},
  2007.

\bibitem{perronnin10cvpr1}
F.~Perronnin, J.~S\'{a}nchez, and Y.~Liu.
\newblock Large-scale image categorization with explicit data embedding.
\newblock In {\em IEEE Conference on Computer Vision and Pattern Recognition},
  2010.

\bibitem{perronnin10eccv}
F.~Perronnin, J.~S\'anchez, and T.~Mensink.
\newblock Improving the {F}isher kernel for large-scale image classification.
\newblock In {\em European Conference on Computer Vision}, 2010.

\bibitem{quattoni09cvpr}
A.~Quattoni and A.~Torralba.
\newblock Recognizing indoor scenes.
\newblock In {\em IEEE Conference on Computer Vision and Pattern Recognition},
  2009.

\bibitem{quelhas05iccv}
P.~Quelhas, F.~Monay, J.-M. Odobez, D.~Gatica-Perez, T.~Tuytelaars, and
  L.~Van-Gool.
\newblock Modeling scenes with local descriptors and latent aspects.
\newblock In {\em International Conference on Computer Vision}, 2005.

\bibitem{rana14accvw}
A.~Rana, J.~Zepeda, and P.~Perez.
\newblock Feature learning for the image retrieval task.
\newblock In {\em Asian Conference on Computer Vision Workshop on Feature and
  Similarity Learning for Computer Vision}, 2014.

\bibitem{razavian14arxiv}
A.~S. Razavian, H.~Azizpour, J.~Sullivan, and S.~Carlsson.
\newblock {CNN} features off-the-shelf: an astounding baseline for recognition.
\newblock {\em arXiv:1403.6382}, Mar. 2014.

\bibitem{sanchez13ijcv}
J.~S\'{a}nchez, F.~Perronnin, T.~Mensink, and J.~Verbeek.
\newblock Image classification with the {F}isher vector: Theory and practice.
\newblock {\em International Journal on Computer Vision}, 105(3):222--245,
  2013.

\bibitem{sanchez15prl}
J.~S\'anchez and J.~Redolfi.
\newblock Exponential family {F}isher vector for image classification.
\newblock {\em Pattern Recognition Letters}, 59:26 -- 32, 2015.

\bibitem{sermanet14iclr}
P.~Sermanet, D.~Eigen, X.~Zhang, M.~Mathieu, R.~Fergus, and Y.~LeCun.
\newblock Overfeat: Integrated recognition, localization and detection using
  convolutional networks.
\newblock In {\em International Conference on Learning Representations}, April
  2014.

\bibitem{simonyan14arxiv}
K.~Simonyan and A.~Zisserman.
\newblock Very deep convolutional networks for large-scale image recognition.
\newblock {\em {Computing Research Repository}}, 1409.1556, 2014.

\bibitem{sivic03iccv}
J.~Sivic and A.~Zisserman.
\newblock {Video Google}: a text retrieval approach to object matching in
  videos.
\newblock In {\em International Conference on Computer Vision}, 2003.

\bibitem{uijlings13ijcv}
J.~Uijlings, K.~van~de Sande, T.~Gevers, and A.~Smeulders.
\newblock Selective search for object recognition.
\newblock {\em International Journal on Computer Vision}, 104(2):154--171,
  2013.

\bibitem{gemert10pami}
J.~van Gemert, C.~Veenman, A.~Smeulders, and J.-M. Geusebroek.
\newblock Visual word ambiguity.
\newblock {\em IEEE Transactions on Pattern Analysis and Machine Intelligence},
  32(7):1271--1283, 2010.

\bibitem{vedaldi10cvpr}
A.~Vedaldi and A.~Zisserman.
\newblock Efficient additive kernels via explicit feature maps.
\newblock In {\em IEEE Conference on Computer Vision and Pattern Recognition},
  2010.

\bibitem{wang08nips}
X.~Wang and E.~Grimson.
\newblock Spatial latent dirichlet allocation.
\newblock In {\em Advances in Neural Information Processing Systems}, 2008.

\bibitem{wei14arxiv}
Y.~Wei, W.~Xia, J.~Huang, B.~Ni, J.~Dong, Y.~Zhao, and S.~Yan.
\newblock {CNN:} single-label to multi-label.
\newblock {\em {Computing Research Repository}}, 1406.5726, 2014.

\bibitem{winn05iccv}
J.~Winn, A.~Criminisi, and T.~Minka.
\newblock Object categorization by learned universal visual dictionary.
\newblock In {\em International Conference on Computer Vision}, 2005.

\bibitem{zhang07ijcv}
J.~Zhang, M.~Marsza{\l}ek, S.~Lazebnik, and C.~Schmid.
\newblock Local features and kernels for classification of texture and object
  categories: a comprehensive study.
\newblock {\em International Journal on Computer Vision}, 73(2):213--238, 2007.

\end{thebibliography}
